\pdfoutput=1

\documentclass[11pt]{article}
\usepackage{acl}

\usepackage{geometry}
\usepackage[utf8]{inputenc}
\usepackage[T1]{fontenc}
\usepackage{microtype}
\usepackage{times}
\usepackage{latexsym}
\usepackage{amssymb}
\usepackage{amsmath}
\usepackage{mathrsfs}
\usepackage{bm}
\usepackage{dsfont}

\usepackage{graphicx}
\usepackage{xcolor}
\usepackage{colortbl}
\definecolor{pastelgreen}{HTML}{97D077}
\definecolor{customisedblue}{HTML}{6C8EBF}
\definecolor{lightblue}{HTML}{DAE8FC}

\usepackage{array}
\usepackage{booktabs}
\usepackage{tabularx}
\usepackage{multirow}
\usepackage{verbatim}
\usepackage{supertabular}
\usepackage{arydshln}
\usepackage{makecell}
\usepackage{tabulary}
\usepackage{threeparttable}

\usepackage{enumitem}  
\usepackage{tcolorbox}
\tcbuselibrary{skins, xparse}

\usepackage{float}
\usepackage{wrapfig}
\usepackage{caption}
\usepackage{subcaption}

\usepackage{CJKutf8}

\usepackage[frozencache,cachedir=minted-cache]{minted}

\usepackage{hyperref}
\usepackage{cleveref}

\usepackage{pgfplots}
\usepackage{pgfplotstable}
\pgfplotsset{compat=newest}
\usepackage{xspace}
\usepackage{ifthen}
\usepackage[normalem]{ulem}
\usepackage{rotating}

\definecolor{citecolor}{RGB}{34,139,34}
\definecolor{lightred}{RGB}{241,140,142}
\definecolor{amber(sae/ece)}{rgb}{1.0, 0.49, 0.0}
\definecolor{battleshipgrey}{rgb}{0.52, 0.52, 0.51}
\definecolor{cadmiumorange}{rgb}{0.93, 0.53, 0.18}
\definecolor{applegreen}{rgb}{0.55, 0.71, 0.0}
\definecolor{cadmiumgreen}{rgb}{0.0, 0.42, 0.24}
\definecolor{forestgreen}{rgb}{0.13, 0.55, 0.13}
\definecolor{red}{rgb}{0.89, 0.0, 0.13}

\newcommand{\tRAG}{\textbf{\texttt{tRAG}}\xspace}
\newcommand{\mRAG}{\textbf{\texttt{monoRAG}}\xspace}
\newcommand{\MRAG}{\textbf{\texttt{MultiRAG}}\xspace}
\newcommand{\cRAG}{\textbf{\texttt{CrossRAG}}\xspace}

\title{Multilingual Retrieval-Augmented Generation \\
for Knowledge-Intensive Task}

\author{
  Leonardo Ranaldi \qquad Barry Haddow \qquad Alexandra Birch\\
  Institute for Language, Cognition and Computation \\
  School of Informatics, University of Edinburgh \\
  10 Crichton Street, Edinburgh\\
  \texttt{\{first\_name.last\_name\}@ed.ac.uk}
}

\begin{document}
\maketitle
\begin{abstract}
Retrieval-augmented generation (RAG) has become a cornerstone of contemporary NLP, enhancing large language models (LLMs) by allowing them to access richer factual contexts through in-context retrieval. While effective in monolingual settings, especially in English, its use in multilingual tasks remains unexplored.

This paper investigates the effectiveness of RAG across multiple languages by proposing novel approaches for multilingual open-domain question-answering. We evaluate the performance of various multilingual RAG strategies, including question-translation (\tRAG), which translates questions into English before retrieval, and Multilingual RAG (\MRAG), where retrieval occurs directly across multiple languages. Our findings reveal that \tRAG, while useful, suffers from limited coverage. In contrast, \MRAG improves efficiency by enabling multilingual retrieval but introduces inconsistencies due to cross-lingual variations in the retrieved content. To address these issues, we propose Crosslingual RAG (\cRAG), a method that translates retrieved documents into a common language (e.g., English) before generating the response. Our experiments show that \cRAG significantly enhances performance on knowledge-intensive tasks, benefiting both high-resource and low-resource languages.
\end{abstract}

\section{Introduction}
\label{sec:intro}

Retrieval-augmented generation (RAG) aims to improve the factuality and memory access of large language models (LLMs) by combining external knowledge during inference \cite{lewis2020retrieval,petroni-etal-2021-kilt}.
RAG is designed to mitigate some of the well-known limitations of LLMs, including the tendency for hallucinations and the lack of specific domain knowledge in the training data \cite{siriwardhana-etal-2023-improving,kandpal2023largelanguagemodelsstruggle,kasai2023realtime,asai-etal-2023-retrieval}.

Augmenting the questions by operating through relevant information retrieved from external corpora, such as
Wikipedia effectively reduced inaccurate generation, thereby notably improving accuracies \cite{gao2024retrievalaugmentedgenerationlargelanguage,fan2024surveyragmeetingllms}.

Nevertheless, previous efforts focused on English as the data language in their experiments, i.e., the language of the user queries and the retrieval corpora. Hence, limited attention is afforded to studying the type and role of non-English queries and retrieving multilingual documents to augment LLMs' capabilities.  To address this gap, \citet{zhang2022makingmiraclmultilingualinformation,thakur2024nomiraclknowingdontknow,li2024bordirlinesdatasetevaluatingcrosslingual} proposed methodologies to evaluate multilingual retrieval.
\citet{chirkova2024retrievalaugmentedgenerationmultilingualsettings,ranaldi2025improvingmultilingualretrievalaugmentedlanguage} analyse of the impact of multilingual documents in the RAG pipeline.

\begin{figure*}[h]
\centering
    \includegraphics[width=0.96\textwidth]{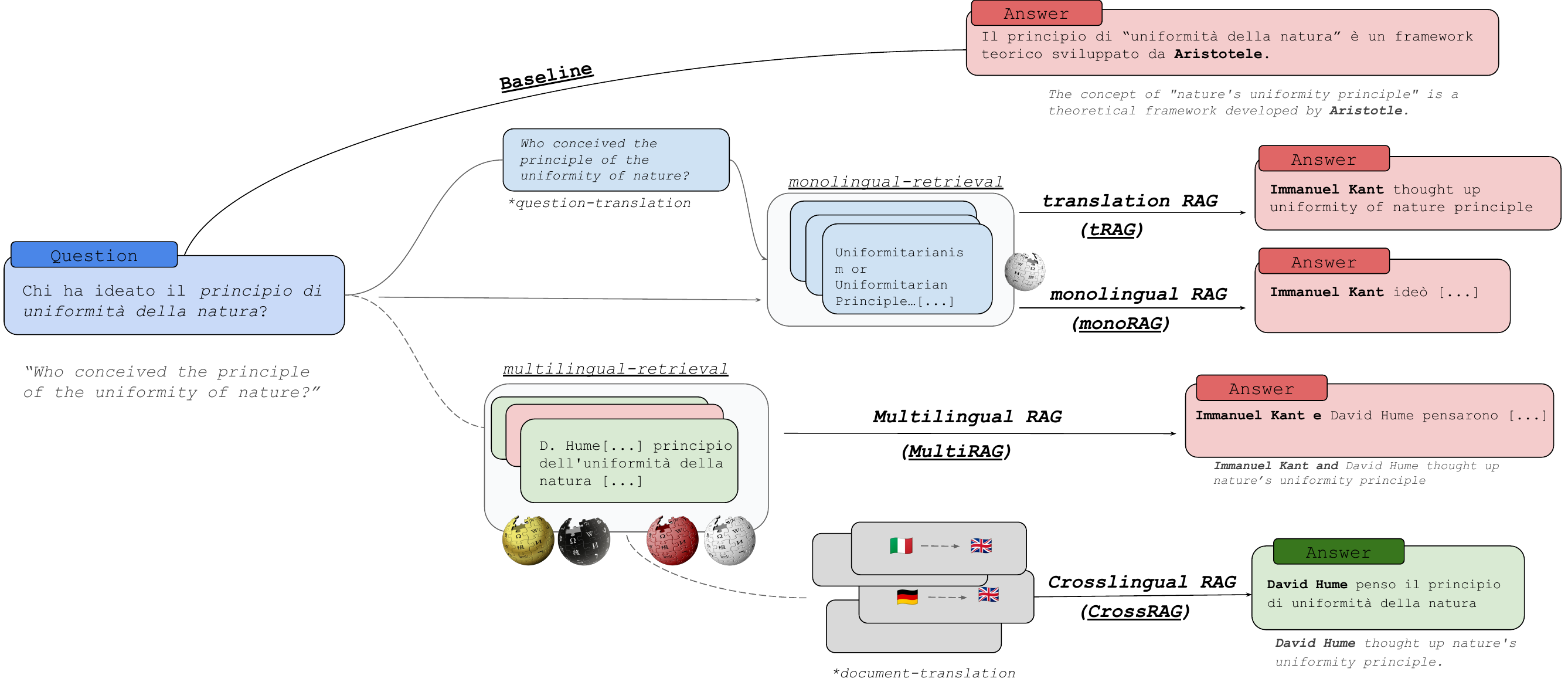}
    \caption{Retrieval-Augmented Generation (RAG) pipelines studied in our works. We explore the performances of different prompting pipelines to handle multilingual queries (\S \ref{sec:methods}). }
    \label{fig:overall_pipeline}
\end{figure*}

In this paper, we systematically investigate the impact of RAG-based pipelines beyond English, aiming to identify potential challenges and propose strategies for improving the performance across a selection of languages. Taking previous work a step further, we evaluate the benefits of extending RAG methodologies in multilingual settings by analysing the effects of different types of retrieved documents on multilingual generative abilities in different languages. 
Complementing the foundation work of \cite{chirkova2024retrievalaugmentedgenerationmultilingualsettings}, we introduce and analyse the trade-off of different approaches that lead LLMs to harness multilingual knowledge.

This leads to the main research questions:

\noindent \textit{RQ1: How does multilingual retrieval affect RAG accuracy and consistency?}\\
\textit{RQ2: What are the benefits and limitations of incorporating multilingual knowledge in RAG models?}\\
\textit{RQ3: Which methods could improve multilingual RAG performance?}\\

To answer these questions, we produce a comprehensive evaluation by introducing strategies for handling multilingual queries as well as retrieval stages, as shown in Figure \ref{fig:overall_pipeline}. 
We use three knowledge-intensive question-answering tasks properly constructed for multilingual evaluation as they best represent multilingual open-ended question-answering tasks \cite{longpre2021mkqalinguisticallydiversebenchmark,chirkova2024retrievalaugmentedgenerationmultilingualsettings}. Then, we employed different LLMs, chosen for proficiency in RAG tasks and multilingual performances, to investigate their capabilities in leveraging multilingual retrieved knowledge.

The main contributions of our paper are:

\begin{itemize}
\item  We explore RAG beyond English by showing the benefits derived from extending the range of retrieval to multilingual contexts. Firstly, we show that naïve approaches such as query translation (\tRAG) generate incorrect translations, leading to wrong retrieval and misleading generations, for instance, Appendix \ref{app:tRAG_wrong}. Instead, Multilingual RAG (\MRAG), based on multilingual retrieval and language-specific queries, outperforms monolingual RAG (\mRAG) based on monolingual retrieval sources in the query language. 

\item We then study the dynamics between the languages that emerge in \MRAG. We outline the advantages of document retrieval over heterogeneous knowledge bases, and at the same time, we display the problems that some models have when they have to operate with retrieved knowledge from documents in different languages. Specifically, we show that the LLMs used in our work are proficient at understanding multilingual questions but fail in extracting information, especially in low-resource languages.

\item In order to address these problems with  \MRAG, we introduce a document-level translation pipeline (\cRAG) that allows the LLMs to handle knowledge-intensive tasks by operating with retrieved documents in a single language (i.e. English) but still providing multilingual responses.

\end{itemize}

\section{Methods}
\label{sec:methods}
Retrieval-augmented Generations (RAG) methods improve performance of LLMs in knowledge-intensive tasks by combining questions with retrieved knowledge in context (\S \ref{sec:RAG}). 

Although the usefulness of RAG has been demonstrated, evaluations and further studies are primarily conducted in English, leaving other languages unexplored. Hence, we propose a systematic study of the portability of RAG pipelines to languages other than English (\S \ref{sec:multilingual-RAG}), analysing different approaches to improve the effective value of expanding retrieval beyond English (\S \ref{sec:in-context-strategies}).

\subsection{RAG Pipelines}
\label{sec:RAG}

In traditional RAG, knowledge is acquired from domains $\mathcal{D}$ (e.g., Wikipedia or internal databases) and used during inference to promote accurate generation. The pipeline is structured into phases: 

\textbf{Retrieval} 
In this phase, the relevant top-$k$ documents $docs = \{d_1, \dots, d_k\}$ are retrieved, based on the query $\mathcal{Q}$ from $\mathcal{D}$, using a retrieval system $\mathcal{R}$.
During retrieval with $\mathcal{R}$, the question $\mathcal{Q}$ and documents in $\mathcal{D}$ are encoded, forming $h_{\mathcal{Q}} = \mathcal{R}(\mathcal{Q}) \in \mathbb{R}^n$ for query and $h_{\mathcal{D}} = \mathcal{R}(\mathcal{D}) \in \mathbb{R}^n$ for documents.
Then, the similarity $\langle h_{\mathcal{Q}}, h_\mathcal{D} \rangle$ is used to select a collection $\mathcal{C}$ from $\mathcal{D}$, consisting of documents that best match the query.
Usually, to improve retrieval quality, the top most relevant documents are filtered and ordered using a re-ranked model obtaining $docs$. Then, they are encoded together with the query in $h_{\mathcal{Q},\mathcal{C}} = \mathcal{R}(\mathcal{Q},\mathcal{C}) \in \mathbb{R}$. 
This allows for $h_{\mathcal{Q},\mathcal{C}}$ representations, which capture similarities between the query and the documents by improving quality and considering similarity, specific contexts, and semantic relevance to $docs = \operatorname{top-k}\{ h_{\mathcal{Q},\mathcal{C}} \}$.

The retriever generally uses a ranker based on architectures trained on specific information retrieval datasets and a customised reranker. 
Since our work focuses on using such systems in a multilingual setting, we operate via retriever and re-ranker systems provided by Cohere\footnote{https://huggingface.co/Cohere/Cohere-embed-multilingual-v3.0} detailed in \S \ref{sec:exp_set}.

\textbf{Inference} The second phase consists of augmenting the LLMs' capabilities in answering a given query $\mathcal{Q}$ using knowledge delivered from reference evidence, i.e. the retrieved relevant documents. The LLM generates the answer $\mathcal{A}$ from $\mathbb{LLM}(\mathcal{Q}, docs)$. 
Here, using a well-defined prompt \textit{template} to get the LLM to consider retrieved documents a source of knowledge is recommended. 
Following the earlier RAG heuristics \cite{gao2024retrievalaugmentedgenerationlargelanguage}, we propose a standard template that instructs the model ("\textbf{\#Instructions}") to consider "\textbf{\#Reference Evidence}" in retrieved documents for delivering the final answer $\mathcal{A}$. An example prompt is reported in Table \ref{tab:table_RAG}.

\begin{table}[]
\begin{tcolorbox}[colback=white,colframe=customisedblue,title=RAG Prompt Template]
\textit{Please answer the question by following the provided instructions.}\\
\textbf{\#Instructions:}\\
\textit{Answer the question as clearly as possible using the provided \textbf{reference evidence} and follow the format 'Answer:'} \\
\textbf{\#Reference Evidence:}\\
\textit{$docs = \operatorname{top-k}\{ h_{\mathcal{Q},\mathcal{C}} \}$, $\mathcal{C} \in \mathcal{D}$}\\
\textbf{\#Question:}\textit{$\mathcal{Q}$}
\end{tcolorbox}
\caption{RAG instructions (prompt) for the model to elicit they to consider the reference evidence (\textit{docs}) for generating an answer to a given question ($\mathcal{Q}$).}
\label{tab:table_RAG}
\end{table}

\subsection{RAG beyond English}
\label{sec:multilingual-RAG}

\paragraph{Monolingual RAG (\mRAG)} In general RAG pipelines (\S \ref{sec:RAG}), it is assumed that the query $\mathcal{Q}$ and $\mathcal{A}$ are in the same language, which we refer to as $L_{SL}$. 
Consequently, the $docs$ retrieved from $\mathcal{D}_{SL}$ are in $L_{SL}$. Hence, in this setting, we instruct the $\mathbb{LLM}$ using the template in Table \ref{tab:table_RAG}, \textit{$docs = \operatorname{top-k}\{h_{\mathcal{Q}_{SL},\mathcal{C}} \}$ where $\mathcal{C} \in \mathcal{D}_{SL}$} and \textit{$\mathcal{Q}_{SL}$} is in $L_{SL}$. For the rest of the paper, we refer to this setting as monolingual RAG (\mRAG). 

\paragraph{Translation RAG (\tRAG)}  Since the knowledge sources from which retrieval is made are generally richer in English \cite{sharma2024fauxpolyglotstudyinformation}, a practical way to solve RAG beyond English is to translate the $\mathcal{Q}_{SL}$ to English $\mathcal{\Tilde{Q}}_{En}$ using a translation system and perform the retrieval from $\mathcal{D}_{En}$. Then, we instruct the $\mathbb{LLM}$ using the same setting of \mRAG, but in contrast, the retrieved documents are $docs=\operatorname{top-k}\{h_{\mathcal{\Tilde{Q}}_{En},\mathcal{C}} \}$ where $\mathcal{C} \in \mathcal{D}_{En}$.

Although these strategies can solve the language barrier by allowing non-English queries to operate, the scope of retrieval is limited to only one domain, namely $\mathcal{D}_{En}$ for \tRAG and $\mathcal{D}_{SL}$ for \mRAG. In addition to the limited scope, the translation also affects retrieval (for instance, see the example in Appendix \ref{app:tRAG_wrong}).

\paragraph{Multilingual RAG (\MRAG)}
Hence, we extend the scope of the retrieval to many languages. We use a retriever whose database consists of $\bigcup_{\mathcal{D}_i \in L}$, i.e.\ the union across all resources in all available languages. As in the \mRAG, we instruct the $\mathbb{LLM}$ to consider the retrieved documents using the template in Table \ref{tab:table_RAG}. In contrast to the previous approaches we use the $docs=\operatorname{top-k}\{ h_{\mathcal{Q}_{SL},\mathcal{C}} \}$, where $\mathcal{C} \in \bigcup_{\mathcal{D}_i \in L}$ and \textit{$\mathcal{Q}_{SL}$}.

\subsection{Cross-lingual RAG}
\label{sec:in-context-strategies}

Multilingual retrieval and prompting strategies broaden the scope of retrieval. As a result, retrieved documents can be in any language in $\mathcal{D}$. 

Although this is a plus for retrieval, it can degrade the LLM's responses, for example, generating answers in the wrong language (see example in Appendix \ref{app:MRAG_bad}) because it must combine in-context documents in different languages.

To solve this issue, we propose Cross-lingual RAG (\cRAG), in which the documents are retrieved as in \MRAG but are then translated by an external tool $\mathbb{T}$ and delivered at inference time in English. This approach improves the accuracy of the response without requiring a substantial additional computational effort.

\section{Experiments}
\label{sec:Experiments}

We select three multilingual open-domain question-answering tasks (\S \ref{sec:dataset}) to compare our approaches. We perform the retrieval and inference phases described in \S \ref{sec:exp_set} and perform the evaluations as presented in \S \ref{sec:evaluation}.

\subsection{Tasks \& Datasets}
\label{sec:dataset}
We use the following question-answering (QA) tasks: \textit{(i)} MLQA \cite{lewis-etal-2020-mlqa}, \textit{(ii)} MKQA \cite{longpre2021mkqalinguisticallydiversebenchmark} and \textit{(iii)} XOR-TyDi QA \cite{asai-etal-2021-xor} as they best represent multilingual open-ended question-answering tasks. These datasets are extensions of resources that originated in English. MLQA and MKQA are manually and machine-translated, whereas XOR-TyDi QA is translated by professional annotators. We provide details about the languages covered and the number of questions in Appendices \ref{app:info_dataset} and \ref{app:languages}.

\subsection{Experimental Setup}
\label{sec:exp_set}

To explore RAG pipelines beyond English, we apply the methods introduced in \S \ref{sec:methods} based on retrieval and inference phases.

\paragraph{Retrieval}
We use Cohere as the retrieval system $\mathcal{R}$ and Wikimedia\_dump as the database $\mathcal{D}$ for all experiments\footnote{This pipeline makes it easy to search Wikipedia for information and to restrict it to specific languages.}. Specifically, in the version\footnote{Cohere/wikipedia-2023-11-embed-multilingual-v3} provided by Cohere, individual articles are embedded with multilingual embedding model \textit{Cohere\_Embed\_V3} (we report the dump composition in  Table \ref{tab:language_distribution_wikimedia_dump}).
Following the approaches proposed by \citet{asai-etal-2023-retrieval,chirkova2024retrievalaugmentedgenerationmultilingualsettings}, we retrieve the most relevant passages and use top$-5$ as in-context knowledge during inference (details in Appendix \ref{app:retrieval}).
As described in \S \ref{sec:multilingual-RAG} we either use \textit{(i)} \textit{monolingual retrieval} (\mRAG and \tRAG) which consists of retrieval on $\mathcal{D}_{SL}$ with documents only in a single specific language, or \textit{(ii)} \textit{multilingual retrieval} (\MRAG and \cRAG) which consists of retrieval on $\bigcup_{\mathcal{D}_i \in L}$ that is the union of multiple $\mathcal{D}_{i}$ in $L$ used in the evaluated task.

\paragraph{Prompting}
We instruct the LLMs using the prompts introduced in \S \ref{sec:methods}. We then include explicit instructions that elicit the model to consider the input query (\textbf{\#Question:}), retrieved documents (\textbf{\#Reference Evidence:}) and deliver the final answer in an \textbf{``evaluated language''} that, by the construction of our experiments, corresponds to the query language.

\paragraph{Translation} As a translation system, in the main discussion, we use \texttt{Google Translate}\footnote{used via Google Translate API python package} to translate for both \tRAG and \cRAG. Furthermore, we investigate the effect of using LLMs as translation systems operating via GPT-4o and other approaches detailed discussed in \S \ref{sec:Results}.

\begin{figure*}[t]
\centering
         \begin{minipage}{0.3\linewidth}
     \centering
     \includegraphics[width=\linewidth]{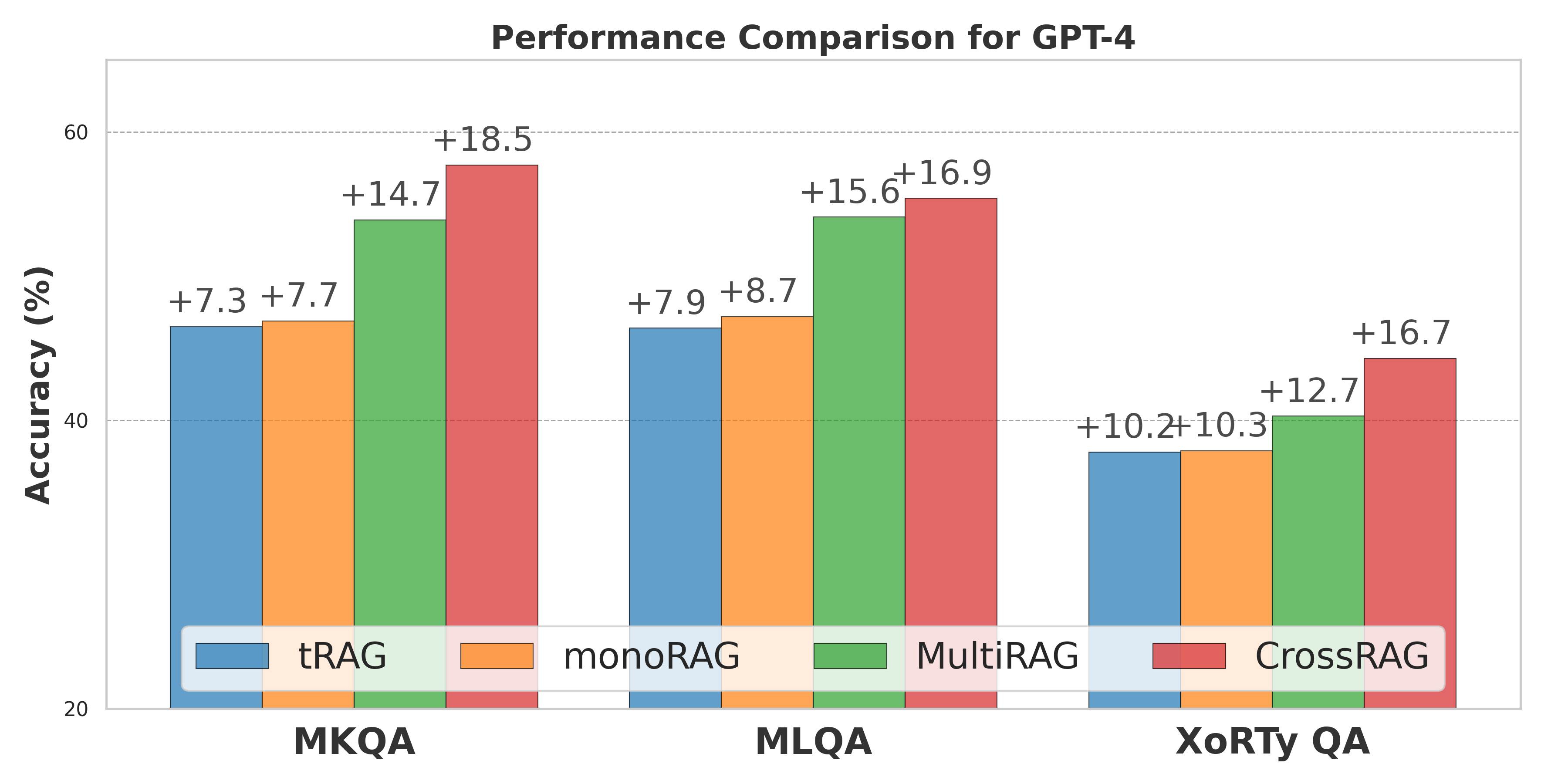}
   \end{minipage}
            \begin{minipage}{0.3\linewidth}
     \centering
     \includegraphics[width=\linewidth]{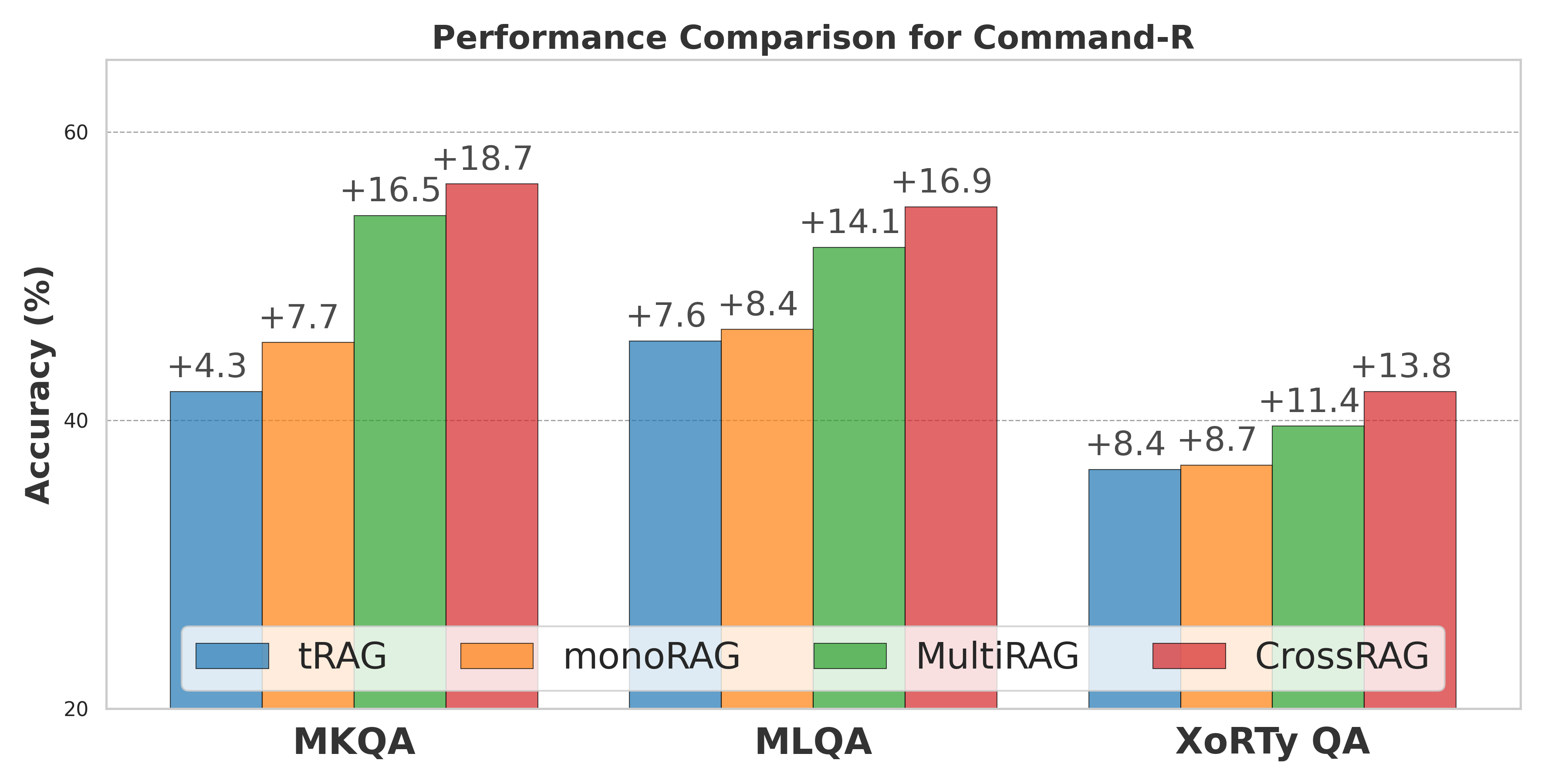}
   \end{minipage}
            \begin{minipage}{0.3\linewidth}
     \centering
     \includegraphics[width=\linewidth]{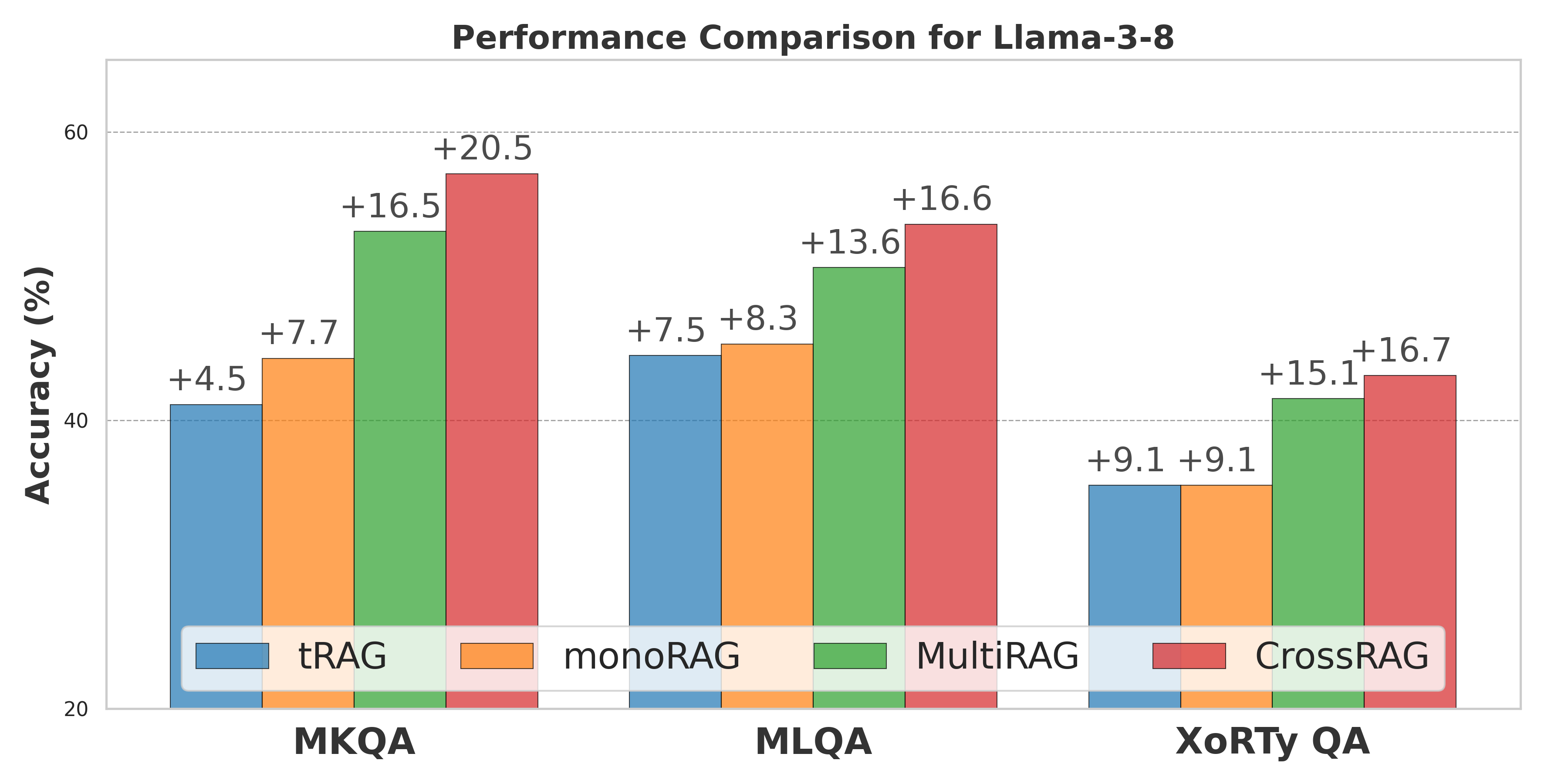}
   \end{minipage}   
   \caption{Performance comparison of models using RAG approaches described in \S \ref{sec:methods} across benchmarks and settings detailed in \S \ref{sec:Experiments}, separated by average (Avg), high-resource (HR) and low-resource (LR) languages averages. The values above the bars are the differences with the baselines (no RAG scores).} 
   \label{fig:results}
\end{figure*}

\paragraph{Models \& Inference Settings}

To get a comprehensive evaluation of existing RAG pipelines, we use three different LLMs: GPT-4o \cite{openai2023gpt4}, Llama-3-8b-instruct \cite{touvron2023llama} and Command-R-35b\footnote{To simplify discussion for the rest of the paper, we will refer to these models using Llama-3-8b and Command-R.} \cite{cohere2024commandr}. Detailed settings and model versions are in Appendix \ref{app:model_versions}.
We use greedy decoding in all experiments to ensure a more deterministic generation process. We set most deterministic temperatures to 0 and the maximum generation length to 2048.

\subsection{Evaluation}
\label{sec:evaluation}
We use flexible exact-match accuracy following \citet{schick2023toolformer,mallen-etal-2023-trust}, which is based on whether or not ground-truth answers are included in the generated answers provided by the models instead of a strict exact match. Furthermore, for a complete comparison, we follow \citet{chirkova2024retrievalaugmentedgenerationmultilingualsettings} to conduct multilingual evaluations using the SQUAD evaluation script and 3-gram character level.

\section{Results \& Discusssions}
\label{sec:Results}

The empirical results across different languages on MKQA, MLQA and XOR TyDi QA are reported in Figure \ref{fig:results}. Overall, the experiments confirm that extending the retrieval scope to multilingual contexts (\MRAG) improves the RAG-based pipelines, outperforming language-specific monolingual RAG (\mRAG) and naïve approaches that address the language barrier using query translation (\tRAG). Indeed, although monolingual retrieval (i.e., \mRAG) achieves benefits compared to the baseline, the retrieved documents may be limited and consequently could not contain the necessary information to answer a language-specific query. Conversely, retrieval from multilingual heterogeneous sources has a broader range of results. However, multilingual knowledge could lead models to wrong generations (see the example in Appendix \ref{app:MRAG_bad}). Therefore, we proposed \cRAG to harness \MRAG retrieval by operating with documents in the same language, i.e. English.   

In the following sections, we analyse the benefits a multilingual retrieval brings when adopted in a RAG strategy (\S \ref{sec:The_impact_RAG_beyond_English}), then we examine the effects across different languages \S \ref{sec:The_Knowledge_Diversity} and propose two strategies to improve the practical usage of the retrieved knowledge in multilingual settings \S \ref{sec:Cross_RAG}. Finally, we conduct additional studies by investigating the role of the retriever and its impact on the final performances (\S \ref{sec:ablation_retriever}), the generated languages (\S \ref{sec:ablation_lan_generated}) and the robustness on challenging perturbations (\S \ref{sec:ablation_robustness}).

\subsection{The impact of RAG beyond English}
\label{sec:The_impact_RAG_beyond_English}

Figure \ref{fig:results} shows the results obtained from different LLMs when prompted with RAG-based strategies in monolingual and multilingual settings as introduced in \S \ref{sec:methods}. 
An overall improvement over baseline models without RAG can be observed using monolingual RAG, i.e., \mRAG (+8.9\% improvement for GPT-4o, +7.9\% improvement for Llama-3-8b and +8.2\% improvement for Command-R). 
Moreover, the results show that the impact of extending retrieval to multilingual settings and using retrieved passages in RAG-based approaches (\MRAG strategy) brings clear benefits. Indeed, performance consistently increases compared to the \mRAG (+5.4\% for GPT-4o, +7.1\% for Llama-3-8b and +5.7\% for Command-R). 
This indicates that multilingual retrieval provides access to broader information that could be unavailable in monolingual resources; for instance, in the example reported in Appendix \ref{app:MRAG_ok}, the information about "England Queens" are not available in Chinese Wikipedia. However, since the scope of retrieval is wider and the retrieval languages are multiple, the languages of the retrieved documents may impact the performance differently, as discussed in \S \ref{sec:The_Knowledge_Diversity}.

\subsection{Knowledge Diversity}
\label{sec:The_Knowledge_Diversity}

\begin{figure}[h]
\centering
    \begin{minipage}{0.9\linewidth}
     \centering
     \includegraphics[width=\linewidth]{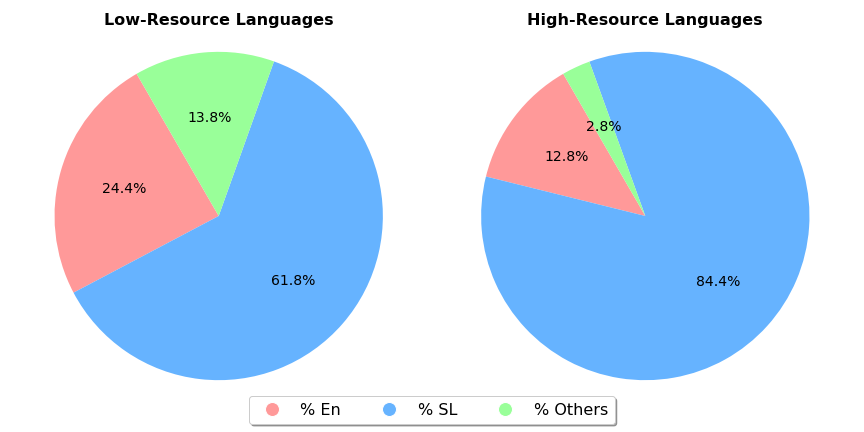}
   \end{minipage} 
      \caption{Average percentage languages of retrieved documents (details in Appendix \ref{tab:retrieved_docs_languages}).} 
   \label{fig:retrieved_docs_languages}
\end{figure}

Multilingual RAG (\MRAG) shows average improvements compared to the baselines, \mRAG and \tRAG, where the retrieved documents are in a single language as discussed in \S \ref{sec:Results}). In \MRAG, the knowledge retrieved are in different languages (as reported in Figure \ref{fig:retrieved_docs_languages}, these are average documents retrieved per language). The differences in percentages are due to the composition of the Wikimedia\_dumps (reported in Table \ref{tab:language_distribution_wikimedia_dump}) undersized for some languages.

\begin{figure}[h]
\centering
    \begin{minipage}{0.9\linewidth}
     \centering
     \includegraphics[width=\linewidth]{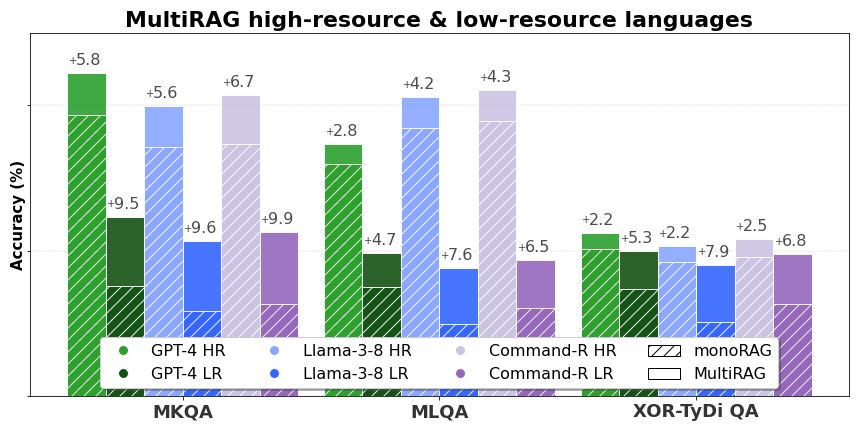}
   \end{minipage} 
      \caption{Accuracies \mRAG and \MRAG in low- (LR) and high-resource (HR) languages. *(differences are above the bars).} 
   \label{fig:performances_mRAG_MRAG}
\end{figure}

Consequently, the effect of \MRAG is different even between languages.
Figure \ref{fig:performances_mRAG_MRAG} shows that the effect of \MRAG on low- (LR) languages is more marked than high-resource (HR) languages\footnote{high- and low-resources defined following \cite{chirkova2024retrievalaugmentedgenerationmultilingualsettings} explained in Appendix \ref{app:low_high_resource_lan}}. 
Indeed, comparing \MRAG with \mRAG in the case of HR, we observe average increases of +3.6\% for GPT-4o, +4.1\% for Llama-3-8b and +4.4\% for Command-R.
In contrast, for LR, there is an average increase of 6.6\% for GPT-4o, 8.4\% for Llama-3-8b and 7.7\% for Command-R).

To gain a comprehensive view of the role of multilingual retrieval conducted in the \MRAG setting, we performed further experiments by restricting the retrieval to a set formed by specific language (SL) and English, which we define as \textit{(En+SL)}. Figure \ref{fig:performances_MRAG_all_en} (Appendix \ref{app:restricted_retrieval_scope}) shows the differences in terms of performance when the scope of the retrieval of \MRAG is broader, i.e. all languages available in the dump used in the experiment (Wikipedia versions as detailed in \S \ref{sec:Experiments}) and \textit{(En+SL)}. On average, extending the scope of retrieval beyond the subset represented of \textit{En+SL} has benefits except GPT-4o in the MKQA task and Llama-3-8b in the case of MLQA.

Hence, although \MRAG consistently achieves higher performance than \mRAG, there are some cases where the heterogeneity of languages is not beneficial. For instance, the case in Appendix \ref{app:MRAG_bad} where passages in English is not taken into account by the model. 
Therefore, to analyse whether the component affecting performances is the ability to leverage the different languages in different retrieved documents, we propose an intervention strategy by introducing a translation phase of the retrieved multilingual knowledge and discuss the results in \S \ref{sec:Cross_RAG}. 

\begin{table}[h!]
\begin{center}
\small
\begin{tabular}{>{\raggedright}p{1.55cm}>{\raggedright}p{1.cm}|ccc|c}
 & & \rotatebox{90}{\textbf{MKQA}} & \rotatebox{90}{\textbf{MLQA}} & \rotatebox{90}{\textbf{XoR Ty-QA}} & \rotatebox{90}{\textbf{Average}} \\ 
\textbf{Model} &  & \rotatebox{90}{$\Delta$} & \rotatebox{90}{$\Delta$} & \rotatebox{90}{$\Delta$} & \\ \hline

\multirow{3}{*}{\textbf{\texttt{GPT-4o}}} 
& \textbf{Avg} & +3.8 & +1.3 & +5.5 & +3.5 \\
& \textbf{HR} & \textbf{+4.2} & +0.7 & +2.7 & +2.5 \\ 
& \textbf{LR} & +1.8 & \textbf{+2.1} & \textbf{+7.1} & \textbf{+3.7} \\ \hline

\multirow{3}{*}{\textbf{\texttt{Command-R}}} 
& \textbf{Avg} & +2.2 & +2.8 & +1.6 & +2.2 \\
& \textbf{HR} & +2.7 & +3.7 & +3.9 & +3.4 \\ 
& \textbf{LR} & \textbf{+5.2} & \textbf{+4.3} & \textbf{+5.5} & \textbf{+5.0} \\ \hline

\multirow{3}{*}{\textbf{\texttt{Llama-3-8b}}} 
& \textbf{Avg} & \textbf{+4.0} & +3.2 & +1.6 & +2.9 \\
& \textbf{HR} & +1.8 & +2.4 & +3.9 & +2.7 \\ 
& \textbf{LR} & +3.7 & \textbf{+4.2} & \textbf{+4.6} & \textbf{+4.1} \\ 
\end{tabular}
\end{center}
\caption{Differences ($\Delta$) between \cRAG and \MRAG.*In \textbf{bold}, the highest differences for model.}
\label{tab:performances_CrossRAGvsMultiRAG}
\end{table}

\subsection{When translating matters}
\label{sec:Cross_RAG}

The red bars in Figure \ref{fig:results} show that the average results obtained by \cRAG are consistently better than those of other approaches. In general, translating the retrieved information into English benefits the final performance. In Table \ref{tab:performances_CrossRAGvsMultiRAG}, we report the performance improvements over \MRAG differentiated for LR and HR. Here, we observe that in HR, there are improvements of around +2.5 for GPT-4o and Llama-3-8b and +3.4 for Command-R when compared to \MRAG. In contrast, we note larger benefits for LR (respectively +3.7 for GPT-4o, +5 for Command-R and +4.1 for Llama-3-8b on average). These results highlight the limitations that the LLMs examined have when operating via \MRAG concerning documents in multiple languages (see the case discussed in \S \ref{sec:The_Knowledge_Diversity} in Appendix \ref{app:MRAG_bad}.

However, since the translation component matters, we proposed the same experimental setting using \textit{(i)} GPT-4o as the translation tool, \textit{(ii)} instruction-tuning at the translation level. 

\begin{table}[h!]
\begin{center}
\small
\begin{tabular}{>{\raggedright}p{1.47cm}>{\raggedright}p{1.7cm}|cc|cc}
 & & \rotatebox{90}{\textbf{\tRAG}} & \rotatebox{90}{\textbf{\tRAG}} & \rotatebox{90}{\textbf{\cRAG}} & \rotatebox{90}{\textbf{\cRAG}} \\ 
\textbf{Model} &  & \rotatebox{90}{\includegraphics[width=1.8em,height=1.7em]{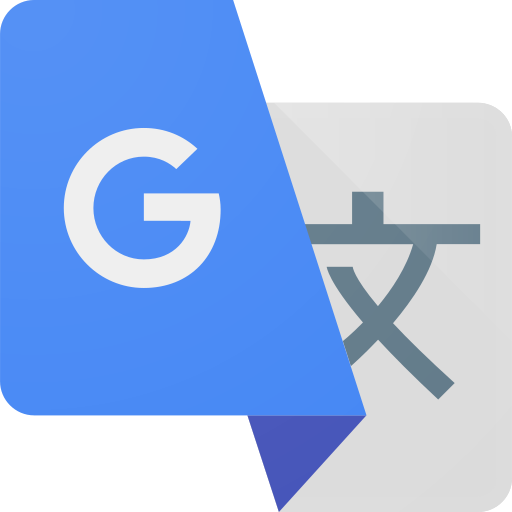}} & \rotatebox{90}{\includegraphics[width=1.5em,height=1.5em]{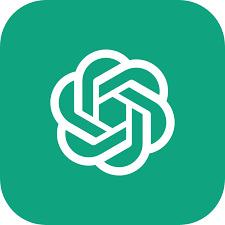}} & \rotatebox{90}{\includegraphics[width=1.5em,height=1.5em]{figures/google_api.png}} & \rotatebox{90}{\includegraphics[width=1.5em,height=1.5em]{figures/gpt.png}} \\ \hline

\multirow{3}{*}{\textbf{\texttt{GPT-4o}}} 
& \textbf{MKQA} & 46.5 & \textbf{48.3} & 60.4 & \textbf{62.0} \\
& \textbf{MLQA} & 46.4 & 47.9 & 55.4 &  \textbf{58.8} \\ 
& \textbf{XoR TDQA} & 37.7 & 38.2 & 45.8 & \textbf{49.3} \\ \hline

\multirow{3}{*}{\textbf{\texttt{Command-R}}} 
& \textbf{MKQA} & 39.9 & 40.3 & 56.4 & 57.8 \\
& \textbf{MLQA} & 45.5 & 46.0 & 54.8 & 56.2 \\ 
& \textbf{XoR TDQA} & 36.6 & 37.3 & 42.0 & \textbf{44.5} \\ \hline

\multirow{3}{*}{\textbf{\texttt{Llama-3-8b}}} 
& \textbf{MKQA} & 41.4 & 42.0 & 57.2 & 58.5 \\
& \textbf{MLQA} & 44.5 & 44.6 & 53.6 & 55.4 \\ 
& \textbf{XoR TDQA} & 37.0 & 37.7 & 44.5 & \textbf{47.3} \\ 

\end{tabular}
\end{center}
\caption{Average performances using two different translation systems. *In \textbf{bold}, the differences that exceed at least 2 points. **XoR TiDy-QA (XoR TDQA)}
\label{tab:avg_results_translation_tool}
\end{table}

\paragraph{GPT-4o as translator}
Here, we propose different settings to observe the effect of various systems on the performance of our \cRAG. Hence, we used GPT-4o (\texttt{GPT-4o} as in \S \ref{sec:exp_set}) as the translation tool. Then, using the prompt in Appendix \ref{app:instruction_template_analysis}, we translated both retrieved documents and questions (in two distinct experimental phases) and reproduced the experimental setting proposed earlier.
Table \ref{tab:avg_results_translation_tool} compares the results using two different systems. In the case of \tRAG, there are no conspicuous improvements (highest difference +1.9 in GPT-4 MKQA). Concerning \cRAG, it can be observed that significant differences emerge between the final results achieved by using Google Translate and GPT-4o. This further demonstrates \textit{(i)} the importance of multilingual retrieval (greater range of retrieval) and \textit{(ii)} the usability of retrieved knowledge by LLM is better when it is in English. In fact, multilingual knowledge retrieved and then processed in English impacts the final generations, whereas the same knowledge (the same documents in a foreign language) does not impact in the same way.

\begin{table}[h]
\small
\center
\begin{tabularx}{0.4\textwidth}{p{1.3cm}<{\centering}p{1cm}<{\centering}p{1cm}<{\centering}p{1cm}<{\centering}}
    \toprule
    \multicolumn{2}{c}{\textbf{Method}} & \textbf{MKQA} & \textbf{MLQA} \\
    \midrule
    \multirow{2}{*}{\MRAG} & \textbf{Avg} & 53.1 & 50.6 \\
                                      & \textbf{LR}  & 44.0 & 38.7 \\
    \midrule
        \multirow{2}{*}{\cRAG} & \textbf{Avg} & 57.2 & 53.8 \\
                                      & \textbf{LR}  & 46.7 & 41.9 \\
    \midrule
    \multirow{2}{*}{\textbf{TF}} & \textbf{Avg} & \textbf{58.9} & 54.7 \\
                                      & \textbf{LR}  & 45.2 & \textbf{43.6} \\
        \midrule
        \multirow{1}{*}{\cRAG} & \textbf{Avg} & 58.5 & \textbf{55.4} \\
                  \textbf{(GPT-4o)}                    & \textbf{LR}  & 47.3 & 42.8 \\
    \bottomrule
\end{tabularx}
\caption{Evaluation using Translation-following (TF), \MRAG and \cRAG (with Google Translate and GPT-4o as translation tools) on Llama-3-8b.}
\label{tab:translation_following}
\end{table}

\paragraph{Translation-following}
Since the language of the retrieved documents plays a crucial role in the model's performance, we propose a multilingual augmentation strategy conceived to enhance their capability to operate multilingual documents. Hence, we employ the Translation-following (TF) approach as proposed in \cite{ranaldi-etal-2024-empowering-multi} and detailed in Appendix \ref{app:Translation-Following}. Table \ref{tab:translation_following} shows that Llama-3-8b enhanced through the TF achieves consistent benefits.
In particular, 11.6\% on MKQA and 7.5\% on MLQA on average values when compared with \MRAG. While 4.9\% on MKQA and 1.8\% on MLQA on average values when compared with \cRAG. Finally, when compared with the \cRAG version with GPT-4o as a translation tool, it achieves comparable performance (differences around <1). However, although performance increases are evident in some cases, there is the cost of additional tuning that should be considered.

\section{Ablation Analysis}
\label{sec:analysis}

The results discussed in \S \ref{sec:Results} show the benefits of \textit{(i)} extending retrieval beyond English contexts and \textit{(ii)} the operability of in-context approaches and translation tools to align the language of different retrieved information. 
This section analyses the qualitative impact of the proposed techniques on generations (\S \ref{sec:ablation_lan_generated}) and the effect of using other retrievals (\S \ref{sec:ablation_retriever}).
Finally, in \S \ref{sec:ablation_robustness}, we study the robustness of LLM to the combination of information in different languages and the number of documents retrieved.

\subsection{Language Generated}
\label{sec:ablation_lan_generated}
\begin{figure}[h!]
\centering
    \begin{minipage}{0.9\linewidth}
     \centering
     \includegraphics[width=0.95\linewidth]{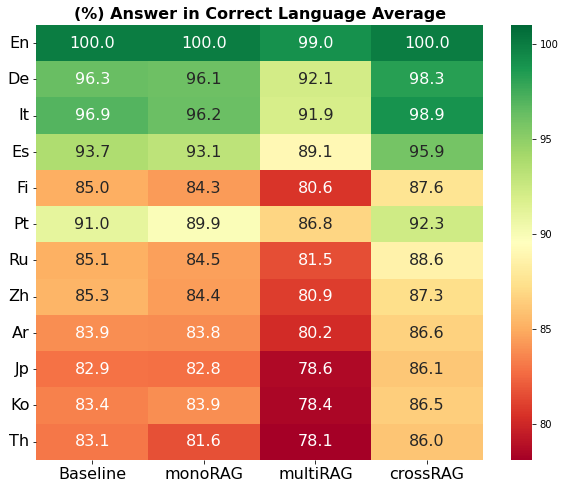}
   \end{minipage} 
      \caption{Average generated languages for MKQA.} 
   \label{fig:generated_languages}
\end{figure}

One of the requirements for the correct answer is that the language must be the same as the query (the labels are also in a specific language). As an evaluation metric, in addition to the accuracy discussed in \S \ref{sec:analysis}, we evaluate the percentage of answers generated in the correct language. To do this, we use the OpenLID framework \cite{burchell-etal-2023-open}. Figure \ref{fig:generated_languages} shows that \cRAG achieves consistently higher rates than \MRAG. Moreover, \mRAG gets comparable performances to the baseline (no RAG approach), but the accuracy is significantly lower. These results demonstrate that the models analysed generally follow instructions (given prompt); however, when operating with multilingual knowledge (i.e., \MRAG), they fail to both follow instructions and deliver the correct response, especially in low-resource languages.  

\subsection{Retrieval Settings}
\label{sec:ablation_retriever}

In our experimental setting (\S \ref{sec:Experiments}), we use Cohere as a retrieval tool. To observe the impact of the retrieval methodologies on the performances, we conducted a parallel experiment using BGE-m3 \cite{chen2024bgem3embeddingmultilingualmultifunctionality} as in \cite{chirkova2024retrievalaugmentedgenerationmultilingualsettings} (detailed in Appendix \ref{app:bergen_retrieval}). Figure \ref{fig:performances_Retriever} shows the average performances obtained by Llama-3-8b on the MKQA subset using the two different retrieval strategies. 
There are no conspicuous differences on average. This indicates that although the retrieval techniques differ, they provide equivalent retrieval methodologies. In our work, we use Cohere because it allows for an already indexed version of Wikipedia dump, as discussed in Appendix \ref{app:retrieval}.

\begin{figure}[h]
\centering
    \begin{minipage}{0.9\linewidth}
     \centering
     \includegraphics[width=\linewidth]{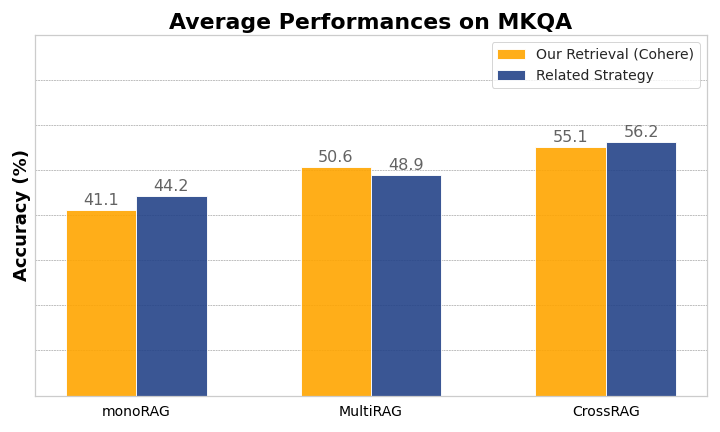}
   \end{minipage} 
      \caption{Difference between our retrieval strategy (\S \ref{sec:Experiments}) and approach proposed in \cite{chirkova2024retrievalaugmentedgenerationmultilingualsettings}} 
   \label{fig:performances_Retriever}
\end{figure}

\subsection{Robustness Analysis}
\label{sec:ablation_robustness}

Figure \ref{fig:robustness_others} shows a robustness analysis of the proposed approaches. We analysed the impact of the order of the retrieved documents by selecting the knowledge provided at the inference phase and conducting an extensive retrieval and a re-ranking (details in \S \ref{sec:exp_set}). To observe the impact of the order of the provided knowledge (documents), we \textit{(i)} randomly shuffled documents (Random Shuffle), \textit{(ii)} English documents in first positions (En doc/s first), \textit{(iii)} English documents at the last positions (En doc/s last). 
From the results in Figure \ref{fig:robustness_others}, it emerges that \cRAG is robust to the varying document order. Instead, \MRAG is more sensitive to retrieval order; this phenomenon emerges in Llama-3-8b and less markedly in Command-R and GPT-4o. 
This further indicates that the language sensitivity of documents is a drawback to the final performance, and operating a translation process or system such as \cRAG improves performance by making it more robust to scenarios where retrieved documents may not be delivered in the most optimal order.

\section{Related Work}

Previous research investigated the advantages of augmenting large language models (LLMs) through retrieved knowledge, a technique known as Retrieval-augmented Generative (RAG) \cite{lewis2020retrieval}.
Many efforts have concentrated on exploring techniques to improve RAG by operating in-context \cite{menick2022teachinglanguagemodelssupport}, tuning \cite{gao-etal-2023-rarr}, or intervening on retrievers \cite{sawarkar2024blendedragimprovingrag}. Although these results represent a considerable step forward, slight attention has been yielded beyond English. 
We work on multilingual tasks involving multilingual queries and documents in the evaluation. While the tremendous effort of \citet{zhang2022makingmiraclmultilingualinformation,thakur2024miragebenchautomaticmultilingualbenchmark} is focused on the study of retrieval from multilingual sources and proposed benchmark-related, we study the role that retrieved documents have on the inference phase of LLMs. Enriching the foundation work proposed by \citet{chirkova2024retrievalaugmentedgenerationmultilingualsettings,sharma2024fauxpolyglotstudyinformation}, we study the impact that different architecture components have on final performance. We analyse the effect of translation at different levels (before and after retrieval) conducted through various tools differing between high and low-resource languages. Analysing criticisms and strengths of Multilingual RAG, we study the roles of different solutions, showing when they lead the LLMs to leverage multilingual knowledge by obtaining consistent benefits.

\section{Conclusion}

RAG has shown great potential in boosting the performance of LLMs on knowledge-intensive tasks. However, scenarios beyond English represent a significant limitation. Hence, we proposed strategies to mitigate these restrictions by introducing retrieval expansion techniques and interventions on retrieved documents. We then analysed the performance of different LLMs in multilingual tasks. The results show that multilingual retrieval brings significant benefits compared to monolingual retrieval or greedy approaches related to query translation. This research shows that a better understanding of RAG-based pipelines beyond English would enable reliable access to information in different languages and cultures. Our contribution supports the need for a strategic combination of the components in RAG pipelines that can aid the performance of various models to complete the LLM perspective in further language landscapes. 

\section*{Acknowledgements}
This work was funded by the European Union’s Horizon Europe (HE) Research and Innovation programme under Grant Agreement No 101070631 (UTTER) and from the UK Research and Innovation (UKRI) under the UK government’s HE funding grant No 10039436.


\bibliography{anthology,custom}

\newpage

\appendix

\appendix

\clearpage

\begin{table}
\section{Models Vesions}
\label{app:model_versions}
\begin{small}
\begin{tabular}{l|p{5.5cm}}
\textbf{Model} & \textbf{Version}  \\ 

\hline
\hline
GPT-4o & OpenAI API (gpt-4-o)  \\
\hline
Command-R & CohereForAI/c4ai-command-r-v01  \\
\hline
Llama3-8b   &  meta-llama/Meta-Llama-3-8B-Instruct \\
\end{tabular}
\end{small}
\caption{List the versions of the models proposed in this work, which can be found on huggingface.co. We used the configurations described in \S \ref{sec:Experiments} in the repositories for each model *(access to the following models was verified on 11 December 2024).}
\label{tab:versions_models}

\section{Difference between High- and Low-resource Languages}
\label{app:low_high_resource_lan}
We define the differences between high-resource (HR) and low-resource (LR) using the consideration already taken in previous works \cite{chirkova2024retrievalaugmentedgenerationmultilingualsettings,ranaldi-etal-2024-empowering-multi}.
Table \ref{tab:language_distribution} reports the language distribution of CommonCrawl, and Table \ref{tab:language_distribution_wikimedia_dump} the number of documents in the Wikipedia dump used in our work (\S \ref{sec:Experiments}).

\begin{center}
\small
\begin{tabular}{|l|c|}
\hline
\textbf{Language} & \textbf{Percentage} \\ \hline
English (en)      & 46.3\%              \\ 
Russian (ru)      & 6.0\%               \\ 
German (de)       & 5.4\%               \\ 
Chinese (zh)      & 5.3\%               \\ 
French (fr)       & 4.4\%               \\ 
Japanese (ja)     & 4.3\%               \\ 
Spanish (es)      & 4.2\%               \\ 
Other             & 23.1\%              \\ \hline
\end{tabular}
\caption{Language distribution of CommonCrawl \cite{commoncrawl2021}.}
\label{tab:language_distribution}
\end{center}

\end{table}

\begin{table}[]
\section{Documents in Wikimedia Dump}

\begin{center}
\small
\begin{tabular}{ll}
\hline

\textbf{Language} & \textbf{Percentage} \\ \hline
English (en)      & 41,488k              \\ 
Russian (ru)      & 13,784k               \\ 
German (de)       & 20,772k               \\ 
Chinese (zh)      & 7,875k               \\ 
Italian  (it)         & 10,462k   \\ 
French (fr)       & 17,813k               \\ 
Japanese (ja)     & 6,626k               \\ 
Spanish (es)      & 12,865k               \\ 
Portuguese (pt)              & 5,637k               \\
Bengali (bn)      & 767k               \\ 
Finnish (fn)          & 272k          \\
Arabic (ar)          &   1,050k      \\
Thai (th)          &      876k   \\
Vietnamese (vi)      & 2,067k \\
Telogu (te)       &  124k \\ \hline

\end{tabular}
\caption{Language distribution of Wikimedia Dump introduced in \S \ref{sec:Experiments}.}
\label{tab:language_distribution_wikimedia_dump}
\end{center}

\end{table}

\begin{table}[t]

\section{Retrieval Details}
\label{app:retrieval}
We use Cohere as the retrieval system and Wikimedia\_dump as the database. Cohere in \textit{wikipedia-2023-11-embed-multilingual-v3} (available on \href{https://huggingface.co/datasets/Cohere/wikipedia-2023-11-embed-multilingual-v3}{huggingface}) provides individual documents embedded with multilingual embedding model \textit{Cohere\_Embed\_V3}. For each question in the evaluation data, we retrieve 50 relevant documents and then rerank the top-5 most relevant ones using dot score between query embedding and document embeddings. We use this procedure as recommended in the use cases (\href{https://huggingface.co/datasets/Cohere/wikipedia-2023-11-embed-multilingual-v3}{case1}, \href{https://github.com/cohere-ai/cohere-developer-experience/tree/main}{case2}).

\section{Retrieval Bergen}
\label{app:bergen_retrieval} 
To reproduce the retrieval setting proposed in \cite{chirkova2024retrievalaugmentedgenerationmultilingualsettings}, we used the open-source library available at the following 
\href{https://github.com/naver/bergen/blob/main/documentation/multilingual.md}{link}. Hence, we reproduce the same settings operating via BGE-m3 \cite{chen2024bgem3embeddingmultilingualmultifunctionality}. 

\end{table}

\begin{table}
\section{Data Composition}
\label{app:info_dataset}
As introduced in \S \ref{sec:dataset} we use \textit{(i)} MLQA \cite{lewis-etal-2020-mlqa}, \textit{(ii)} MKQA \cite{longpre2021mkqalinguisticallydiversebenchmark} and \textit{(iii)} XOR-TyDi QA \cite{asai-etal-2021-xor} as they best represent multilingual open-ended question-answering tasks. MLQA is manually translated from SQuAD v1.1 \cite{rajpurkar-etal-2016-squad}, MKQA and XOR-TyDi QA are machine translated and manually controlled by Natural Questions \cite{kwiatkowski-etal-2019-natural} and TyDi QA \cite{clark-etal-2020-tydi}, respectively. We use parts of the datasets in the languages listed in Table \ref{tab:multilingual_datasets}. For each language, we used the same questions and, consequently, the same number of questions to avoid any imbalance in double-checking by retrieving the corresponding ids. Details on the number of instances are given in Table \ref{tab:num_instrances}.

\hspace{1em}
\begin{small}
    
\begin{tabular}{l|ccc}
\textbf{Dataset} & \textbf{\#language} & \textbf{\#language} & \textbf{\#Total} \\ 
 & \textbf{available} & \textbf{used} & \textbf{used} \\ 

\hline
\hline
MLQA & 1.5$k$ & 0.8$k$ & 9.6$k$ \\
MKQA & 2$k$ & 1.2$k$ & 8.4$k$ \\
XOR-TyDi QA & 0.6$k$ & 0.4$k$ & 2.4$k$ \\
\end{tabular}
\end{small}
\caption{Number of instances used in our evaluations equally distributed among the languages in Table \ref{tab:multilingual_datasets}. We denote by $k$ 1000 instances.}
\label{tab:num_instrances}

\end{table}

\begin{table}[]
\section{Translation-Following}
\label{app:Translation-Following}
We instruct Llama-3-8b using instruction set composed from \textbf{Instruction}: \texttt{‘Translate the following text from $L_X$ to English’}, \textbf{Input:} \texttt{‘Sentence in $L_X$’.} and \textbf{Output:} \texttt{‘Sentence in English’.}, as in \cite{ranaldi-pucci-2023-english} and later \cite{ranaldi-etal-2024-empowering-multi}. We used \textit{news\_commentary} \cite{tiedemann-2012-parallel} by selecting En-X translations as detailed in Table \ref{tab:Translation-Following_Languages}. We randomly extracted 2k demonstrations for the available languages in the open repo (\href{https://huggingface.co/datasets/wecover/OPUS_News-Commentary}{link1}, \href{https://huggingface.co/datasets/Helsinki-NLP/news_commentary}{link2}).
We tune the model for three epochs with a batch size of 32 and a learning rate equal to 1e-5 with a 0.001 weight decay. We use the cosine learning rate scheduler with a warmup ratio of 0.03. We conducted our experiments on a workstation with four Nvidia RTX A6000 48VRAM for approximately 14 GPU/h.

\hspace{0.2cm}
\begin{center}
\begin{small}
\begin{tabular}{p{7cm}} 
\toprule
 \textbf{Languages}  \\ \midrule
German, Spanish, Italian, Russian, Chinese, Japanese, Arabic, Russian, Hindi    \\  \hline 
Total: 18k \\
        
\bottomrule
\end{tabular}
\caption{Instances and languages used for conducting Translation-following experiment.}
\label{tab:Translation-Following_Languages}
\end{small}
\end{center}

\end{table}

\begin{table}[]
\section{Translation-Following Results}
\label{app:Translation-Following_languages}
To make the experimental setting fair and consistent, we randomly extracted the same number of demonstrations for the languages available in the onen-source repositories. Although these offer a large amount of languages, some languages (Korean, Finnish, Thai and Vietnamese) are not available. However, we chose not to exclude these languages from the final evaluation.

\begin{center}
\small
\begin{tabular}{lll}
\hline
\textbf{Language} & \textbf{MLQA} &  \textbf{MKQA} \\ \hline 
German      & 69.8   & 67.5 \\ 
Italian       & 69.2 & - \\ 
Chinese      &  64.4   & 62.3 \\ 
Japanese     & 56.8  &  - \\ 
Spanish      & 69.2   &  69.6 \\ 
Portuguese   & 69.0 & - \\
Russian     & 64.2 & - \\
Arabic      & 59.1 & 43.9 \\
Hindi      & - & 40.6 \\
Finnish  & 57.0 & - \\
Korean  & 39.4 & - \\
Thai  & 23.9 & - \\
Vietnamese  & - & 44.5 \\
 \hline
\textbf{Avg} & 58.9 & 54.7 \\
\textbf{Avg LR} & 45.2 & 43.3 \\
\bottomrule

\end{tabular}
\caption{Performances Llama-3-8b with Translation-following tuning.}
\label{tab:Translation-Following_results}
\end{center}

\end{table}

\begin{table*}[t]
\section{Proposed Task}
\label{app:languages}
\centering
\tiny
\begin{tabular}{cllc}
\hline
\textbf{Dataset}    & \textbf{Task}               & \textbf{Languages} & \textbf{\#Languages} \\ \hline

\textbf{MKQA}      & QA  & \texttt{English, Spanish, German, Italian,} & \\
        &   & \texttt{Portuguese, Russian, Chinese, } &  12\\ 
        &   & \texttt{Korean, Thai, Japanese,} & \\ 
                &   & \texttt{Finnish, Arabic} & \\ 
                \hdashline

   \textbf{MLQA}       & QA  & \texttt{English, Chinese, Arabic, German,} & 7 \\
        &   & \texttt{Spanish, Vietnamese, Hindi,} & \\ 
        \hdashline

    \textbf{XORTyDi}    & QA  & \texttt{English, Chinese, Arabic, Chinese,} & 8 \\
        &   & \texttt{Korean, Finnish, Telogu,} & \\ 
                &   & \texttt{Bengali} & \\

 \hline

\end{tabular}
\caption{Languages present in datasets used in this work. *We denote question-answering task as (QA)}
\label{tab:multilingual_datasets}

\end{table*}

\begin{figure*}[t]
\section{Robustness Analysis}
\label{app:Robustness}
\centering

         \begin{minipage}{0.3\linewidth}
     \centering
     \includegraphics[width=\linewidth]{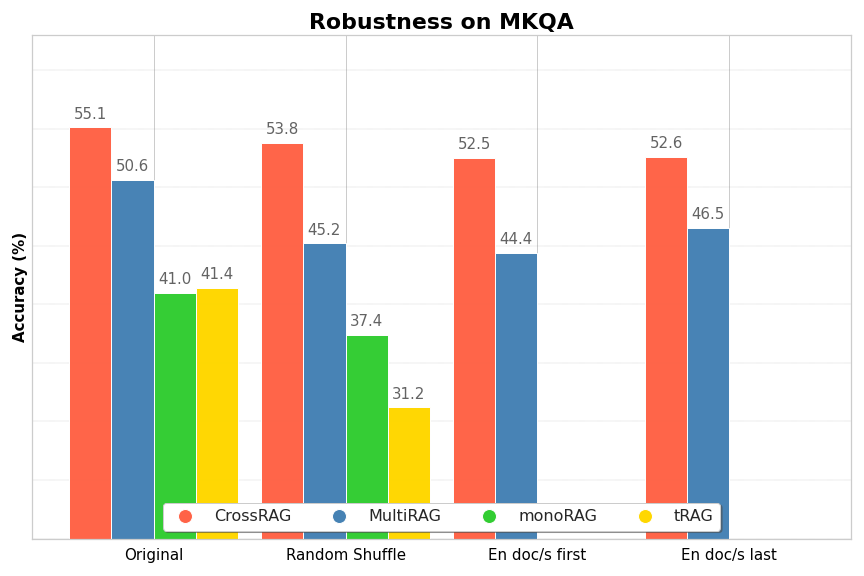}
   \end{minipage}
            \begin{minipage}{0.3\linewidth}
     \centering
     \includegraphics[width=\linewidth]{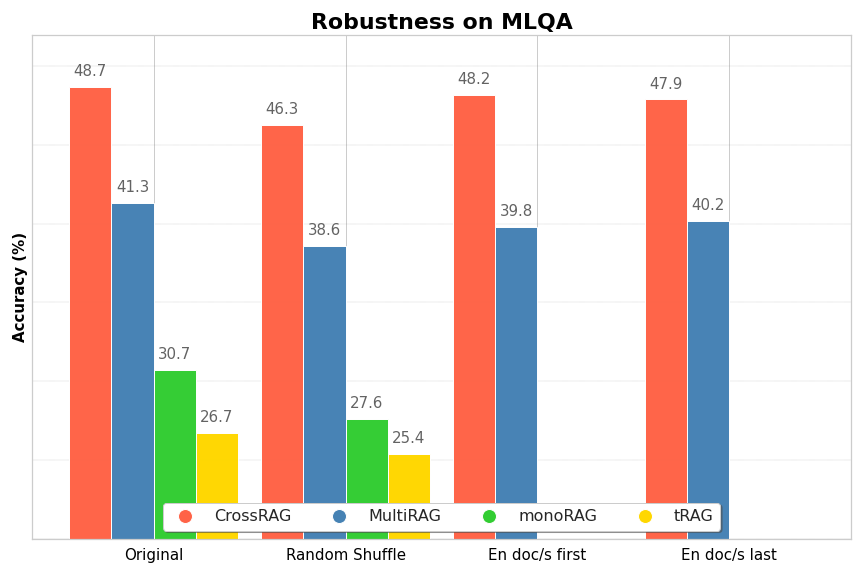}
   \end{minipage}
            \begin{minipage}{0.3\linewidth}
     \centering
     \includegraphics[width=\linewidth]{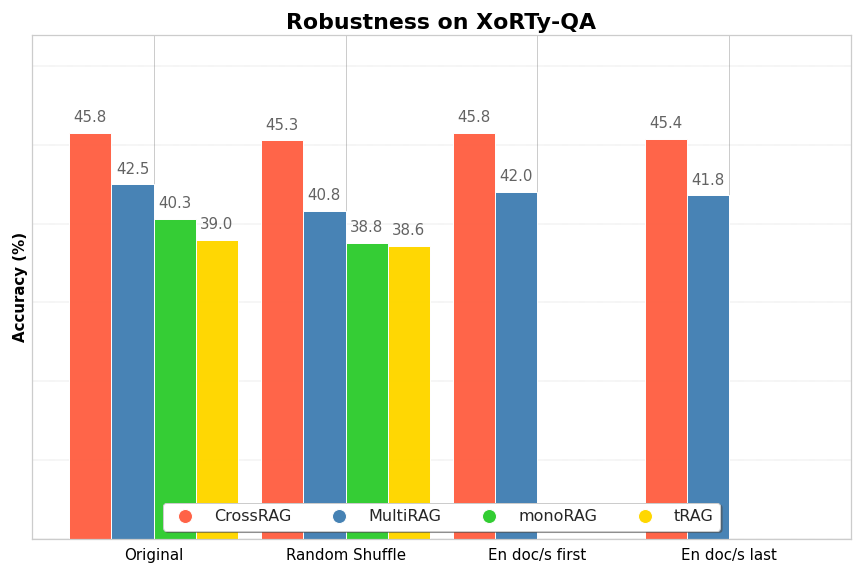}
   \end{minipage}

         \begin{minipage}{0.3\linewidth}
     \centering
     \includegraphics[width=\linewidth]{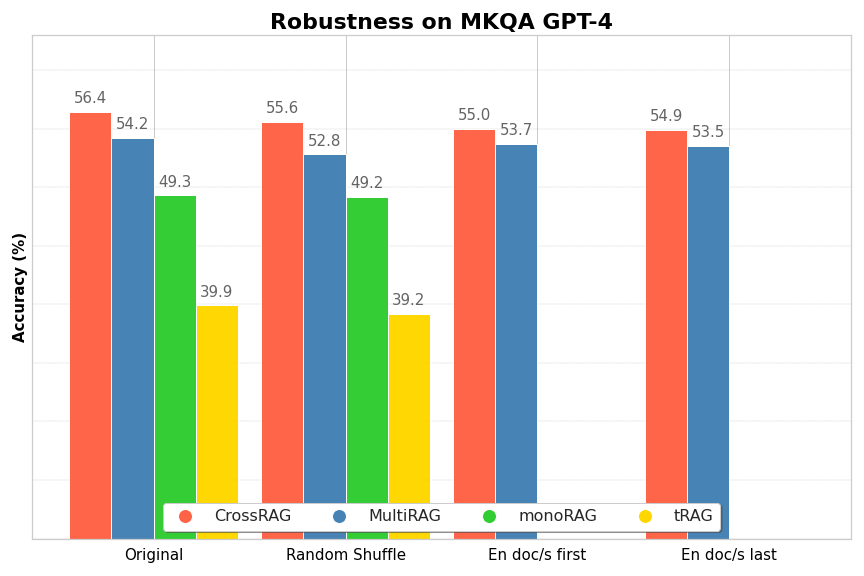}
   \end{minipage}
            \begin{minipage}{0.3\linewidth}
     \centering
     \includegraphics[width=\linewidth]{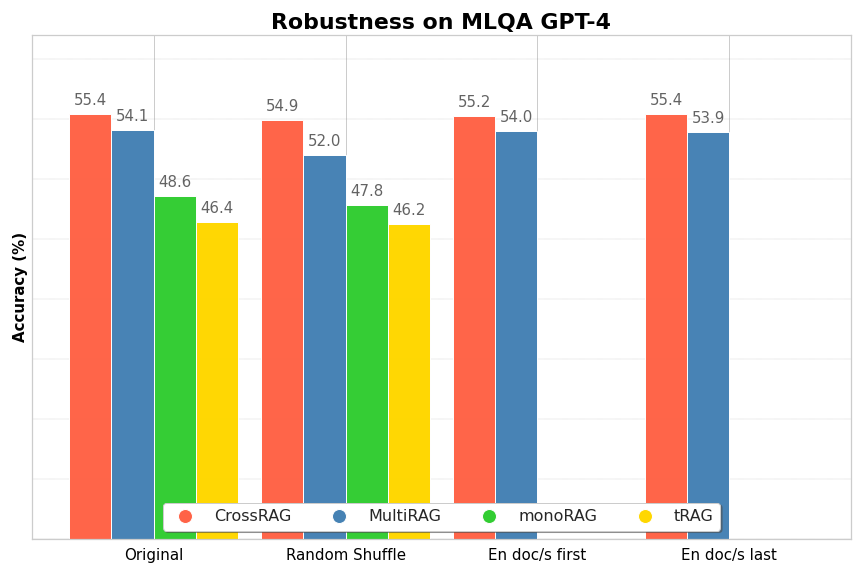}
   \end{minipage}
            \begin{minipage}{0.3\linewidth}
     \centering
     \includegraphics[width=\linewidth]{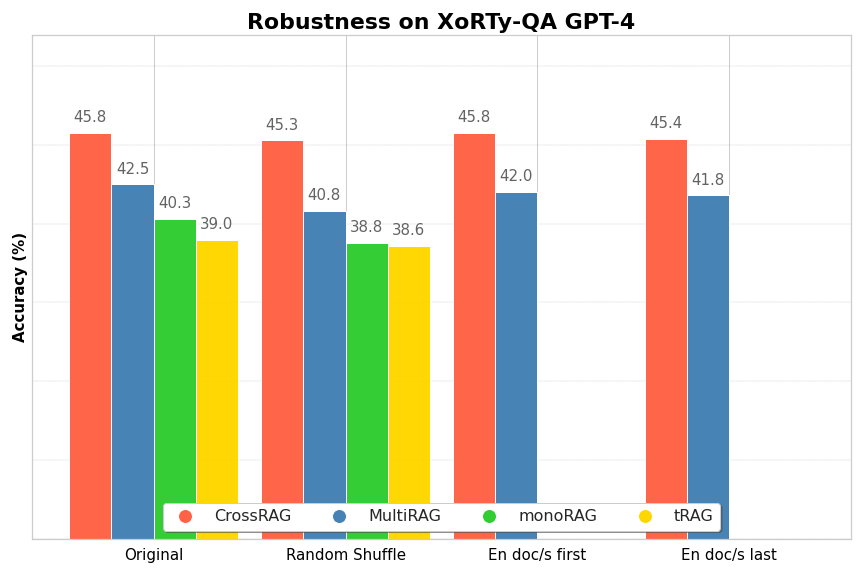}
   \end{minipage}

            \begin{minipage}{0.3\linewidth}
     \centering
     \includegraphics[width=\linewidth]{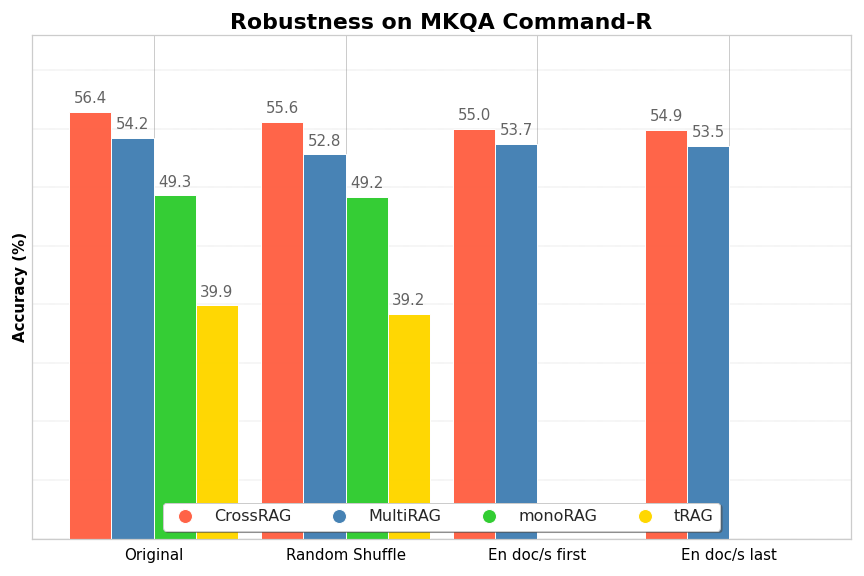}
   \end{minipage}
            \begin{minipage}{0.3\linewidth}
     \centering
     \includegraphics[width=\linewidth]{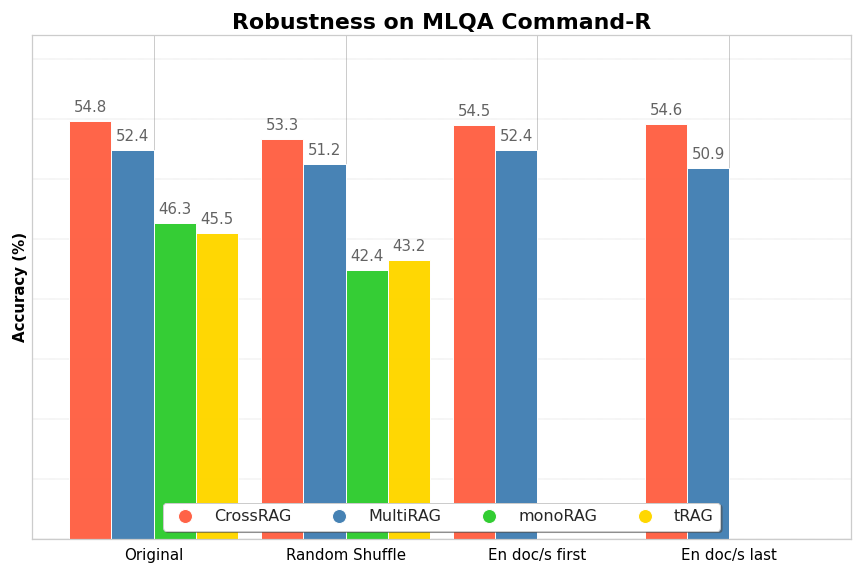}
   \end{minipage}
            \begin{minipage}{0.3\linewidth}
     \centering
     \includegraphics[width=\linewidth]{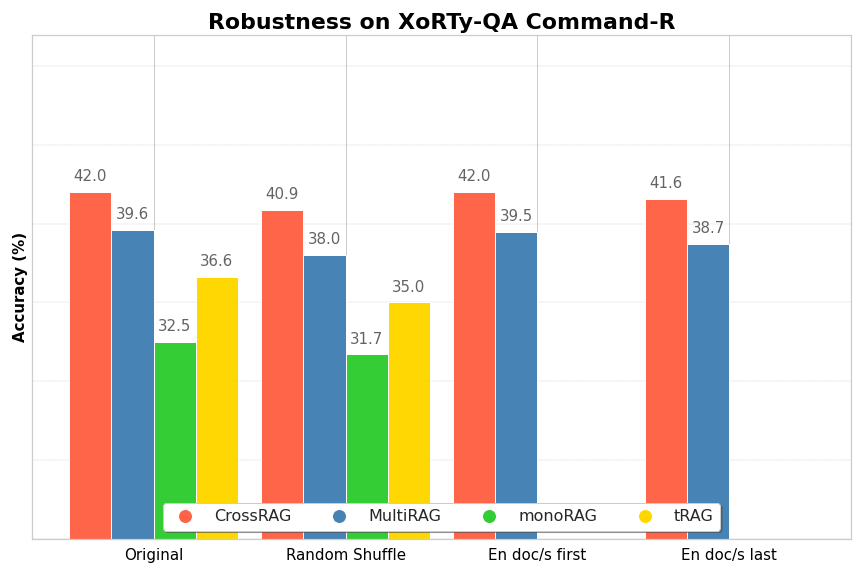}
   \end{minipage}
   
   \caption{Robustness analysis. We deliver the retrieved documents using the order presented in \S \ref{sec:exp_set} (Original), randomly (Random Shuffle), with English first (En doc/s first) and English last (En doc/s last).} 
   \label{fig:robustness_others}
\end{figure*}

\begin{table*}
\section{Instruction Template}
\label{app:instruction_template_analysis}
This section contains the \textit{Instruction Templates} used for the additional analysis.
\begin{center}
\begin{tabular}{|p{1.\linewidth}|}
\hline
\rowcolor{gray!50}
\texttt{\uline{\textbf{Translation}}} \\
Please answer the question by following the provided instructions. \\
\textbf{\#Instructions:}\\
Provide the English translation for this document. Your language and style should align with the language conventions of a native speaker. \\
\textbf{\# Document:} \textit{[document]} \\
\\
\hline
\end{tabular}

\caption{\textit{Instruction Templates}. The structure is defined by a set of in-context examples (zero examples, in the 0-shot case), the question in  \{\textbf{evaluated language}\}, the final instruction part and a special template to guide generation and support the final evaluation.}
\label{tab:task_template}
\end{center}

\end{table*}

\begin{table*}[ht]
\section{Languages of Retrieved Documents}
\label{app:Languages_of_Retrieved_Documents}
\centering
\small
\begin{tabular}{cl|rc|ccc}
\hline
& & \multicolumn{2}{|>{\columncolor[gray]{.95}}c|}{\textbf{\texttt{Retrieval from W$_{En+SL}$}}}  & \multicolumn{3}{|>{\columncolor[gray]{.95}}c}{\textbf{\texttt{Retrieval from W$_{ALL}$}}} \\

&  & \multicolumn{2}{|>{\columncolor[gray]{.95}}c|}{\centering {\textit{($\mathcal{R}$ from En and SL Docs)}}}  & \multicolumn{3}{|>{\columncolor[gray]{.95}}c}{\textit{($\mathcal{R}$ from All Available Docs (ALL))}} \\
  \hline
& \textbf{Question Language} & \textbf{\% En} & \textbf{\% SL} & \textbf{\% En} & \textbf{\% SL} & \textbf{\% Others} \\
\hline
\hline

\multirow{12}{*}{\textbf{\texttt{MKQA}}} &
English & - & - & 98.9\% & - & 1.1\%  \\
& German & 10.8\% & 89.2\% & 10.2\% & 86.3\% & 3.1\%  \\
& Italian  & 12.6\% & 87.4\% & 11.8\% & 85.8\% & 2.4\%  \\
& Spanish  & 12.4\% & 87.6\% & 11.4\% & 86.0\% & 2.8\%  \\
& \uline{Finnish}  & \textbf{26.3\%} & 73.7\% & 22.6\% & 67.1\% &  10.3\% \\
& Portuguese  & 12.0\% & 88.0\% & 11.7\% & 85.8\% & 2.9\%  \\
& Russian  & \textbf{25.3\%} & 74.7\% & 22.2\% & 65.2\% &  12.6\% \\
& Chinese  & 16.3\% & 83.7\% & 14.4\% & 81.2\% & 4.4\%  \\
& \uline{Arabic}  & \textbf{28.2\%} & 71.8\% & 24.3\% & 66.2\% &  9.5\% \\
& Japanese  & 18.2\% & 81.8\% & 16.3\% & 80.3\% & 5.4\%  \\
& \uline{Korean}  & \textbf{30.0\%} & 70.0\% & 24.0\% & 65.5\% &  10.5\% \\
& \uline{Thai}  & \textbf{33.3\%} & 66.7\% & 26.2\% & 64.6\% &  9.2\% \\
\hline
\hline
\multirow{7}{*}{\textbf{\texttt{MLQA}}} &
English & - & - & 99.2\% & - & 0.8\%  \\
& Chinese  & 18.4\% & 82.6\% & 15.3\% & 83.5\% & 2.2\%  \\
& \uline{Arabic}  & \textbf{28.1\%} & 71.9\% & 20.8\% & 70.0\% &  9.2\% \\
& German & 14.4\% & 85.6\% & 13.0\% & 85.5\% & 1.5\%  \\
& Spanish  & 10.7\% & 89.3\% & 11.4\% & 86.0\% & 2.8\%  \\
& \uline{Vietnamese} & \textbf{39.0\%} & 61.0\% & 32.2\% & 55.4\% &  12.4\% \\
& \uline{Hindi} & \textbf{38.5\%} & 61.5\% & 32.6\% & 58.8\% &  9.2\% \\
\hline
\hline
\multirow{7}{*}{\textbf{\texttt{XORTyDi QA}}} &
English & - & - & 98.4\% & - & 1.6\%  \\
& Arabic  & 18.4\% & 81.6\% & 16.3\% & 76.6\% &  7.1\% \\
& Bengali & 43.8\% & 56.2\% & 40.6\% & 46.6\% &  12.8\% \\
& Chinese  & 16.8\% & 83.2\% & 15.6\% & 79.0\% & 7.4\%  \\
& Korean  & 34.3\% & 65.7\% & 31.2\% & 59.2\% &  9.8\% \\
& Russian  & 23.6\% & 76.4\% & 19.8\% & 68.4\% &  11.8\% \\
& Finnish  & 20.6\% & 79.4\% & 19.8\% & 70.8\% &  9.4\% \\
& Telogu & 45.6\% & 54.4\% & 42.0\% & 45.6\% &  12.4\% \\
\hline
\end{tabular}
\caption{Percentage of the languages of retrieved documents. We retrieve the documents using $\mathcal{R}$ system from the Wikipedia dump (detailed in \S \ref{sec:Experiments}) considering both English+Specific Language (\textbf{\texttt{W$_{En+SL}$}}) and all languages analysed in the task (\textbf{\texttt{W$_{ALL}$}}). The languages are checked using OpenLID framework \cite{burchell-etal-2023-open}.}
\label{tab:retrieved_docs_languages}
\end{table*}

\begin{figure*}[h]
\section{Performance differences between Retrieval Scope}
\label{app:restricted_retrieval_scope}
\begin{center}
    \begin{minipage}{0.44\linewidth}
        \centering
        \includegraphics[width=\linewidth]{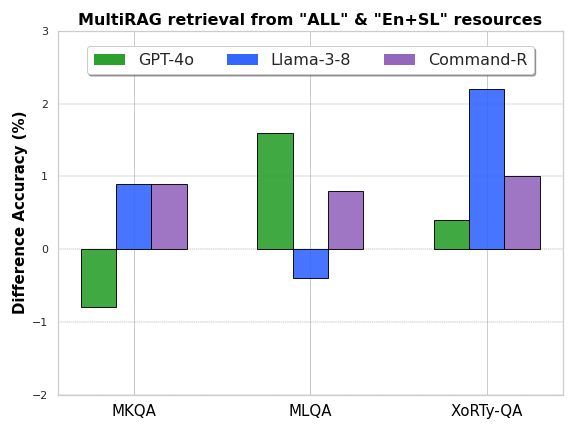}
        \caption*{(a) MRAG Differences (ALL vs En+SL)}
    \end{minipage}
    \begin{minipage}{0.44\linewidth}
        \centering
        \includegraphics[width=\linewidth]{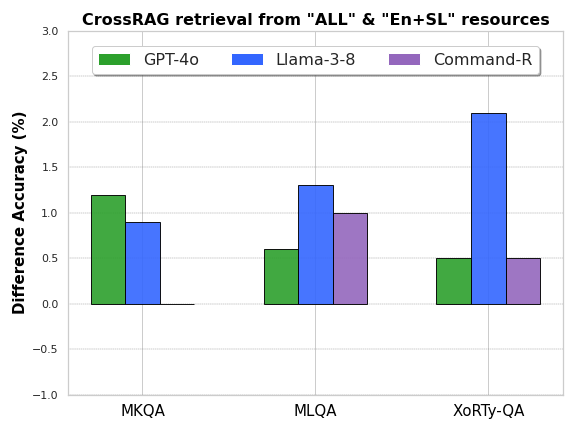}
        \caption*{(b) CRAG Differences (ALL vs En+SL)}
    \end{minipage}
    \end{center}
    \caption{Difference in MRAG (a) and CRAG (b) between retrieval from documents available in Wikidump (ALL) and retrieval done in English+Query specific language (En+SL).}
    \label{fig:performances_MRAG_all_en}
\end{figure*}

\begin{table*}[h]
\section{Perfomances using character 3-gram}
\centering
\tiny
\resizebox{0.9\textwidth}{!}{
\begin{tabular}{lccccccccc}
\toprule
\multirow{2}{*}{\textbf{Model}} & \multicolumn{3}{c}{\textbf{MKQA}} & \multicolumn{3}{c}{\textbf{MLQA}} & \multicolumn{3}{c}{\textbf{XoRTy-QA}} \\
\cmidrule(lr){2-4} \cmidrule(lr){5-7} \cmidrule(lr){8-10}
& \textbf{Avg} & \textbf{HR} & \textbf{LR} & \textbf{Avg} & \textbf{HR} & \textbf{LR} & \textbf{Avg} & \textbf{HR} & \textbf{LR} \\
\midrule
\midrule
GPT-4-o & 38.3 & 55.5 & 30.3 & 41.2 & 50.2 & 30.1 & 29.8 & 35.2 & 27.4 \\
\tRAG & 50.2 & 61.1 & 38.3 & 50.4 & 58.5 & 39.5 & 38.6 & 41.3 & 40.5 \\
\mRAG & 49.5 & 60.5 & 38.0 & 50.4 & 60.5 & 37.7 & 39.7 & 42.1 & 37.6 \\
\MRAG & 55.7 & 63.7 & 45.4 & 55.9 & 66.4 & 46.5 & 42.1 & 44.4 & 40.5 \\
\cRAG & 58.0 & 66.6 & 47.9 & 58.6 & 68.6 & 48.6 & 44.7 & 46.5 & 43.2 \\
\midrule
\midrule
Command-R & 40.2 & 48.7 & 31.8 & 40.1 & 50.5 & 30.0 & 31.0 & 34.5 & 24.9 \\
\tRAG & 42.1 & 53.8 & 31.0 & 47.8 & 59.8 & 36.1 & 38.9 & 40.9 & 37.5 \\
\mRAG & 46.7 & 54.9 & 35.1 & 48.6 & 60.5 & 37.1 & 35.0 & 43.6 & 27.1 \\
\MRAG & 57.0 & 63.9 & 45.8 & 54.5 & 64.3 & 45.3 & 42.0 & 44.3 & 41.9 \\
\cRAG & 59.3 & 67.0 & 51.1 & 57.2 & 68.1 & 46.1 & 44.4 & 48.4 & 44.5 \\
\midrule
Llama-3-8 & 38.7 & 47.4 & 30.9 & 39.9 & 49.2 & 29.1 & 28.9 & 32.2 & 26.3 \\
\tRAG & 43.9 & 52.2 & 29.5 & 47.3 & 58.8 & 34.8 & 37.9 & 39.2 & 36.3 \\
\mRAG & 43.5 & 54.0 & 33.5 & 47.8 & 59.4 & 41.6 & 39.9 & 41.1 & 36.5 \\
\MRAG & 53.5 & 62.1 & 43.6 & 52.7 & 63.8 & 40.5 & 44.1 & 42.9 & 38.2 \\
\cRAG & 57.4 & 65.3 & 51.0 & 56.2 & 66.9 & 45.8 & 45.7 & 46.8 & 44.2 \\
\midrule
\bottomrule
\end{tabular}
}
\caption{Performance (\textit{character 3-gram recall} as in \cite{chirkova2024retrievalaugmentedgenerationmultilingualsettings}) using RAG approaches described in \S \ref{sec:methods} across benchmarks and settings detailed in \S \ref{sec:Experiments}, separated by total average (Avg), high-resource (HR) and low-resource (LR) languages averages.}
\label{tab:results_3-gram}
\end{table*}

\begin{table*}[h!]
\section{Example of bad retrieval in \tRAG}
\label{app:tRAG_wrong}
\centering
\small
\begin{tabular}{p{16cm}}
\textbf{Original Question:} \textit{¿quién escribió variaciones de Campanita del lugar?} \\
\textbf{Translated Question:} \textit{Who wrote variations of Tinkerbell of the Place?} *\uline{translation by Google API} \\

\textbf{Target:} \textit{[Wolfgang Amadeus Mozart, Mozart]} \\
\midrule

\textbf{[1]:} Tinker Bell is a fictional character from J. M. Barrie's 1904 play Peter Pan and his 1911 novelisation \textit{Peter and Wendy}. She has appeared in a variety of film and television adaptations of the Peter Pan stories, notably Walt Disney's 1953 animated film \textit{Peter Pan} and its 2023 live-action adaptation \textit{Peter Pan \& Wendy}. \\

\textbf{[2]:} ‘Jingle Bells’ is one of the best known and most sung traditional winter songs in the world. It was written between 1850 and 1857 by the American composer James Pierpont (1822– 893) under the title ‘The One Horse Open Sleigh’ and was published in Boston by Oliver Ditson \& Co. on 16 September 1857. \\

\textbf{[3]:} Tinker Bell is a 2008 American animated film and the first installment in the Disney Fairies franchise produced by DisneyToon Studios. It is about Tinker Bell, a fairy character created by J. M. Barrie in his 1904 play \textit{Peter Pan, or The Boy Who Wouldn't Grow Up}, and featured in the 1953 Disney animated film, \textit{Peter Pan} and its 2002 sequel \textit{Return to Neverland}. Unlike Disney’s two Peter Pan films featuring the character, which were produced primarily using traditional animation, Tinker Bell was produced using digital 3D modeling. The film was released on DVD and Blu-ray by Walt Disney Studios Home Entertainment on October 28, 2008. A sequel, \textit{Tinker Bell and the Lost Treasure}, was released in 2009. \\

\textbf{[4]:} Tinker Bell is an animated fantasy film series based on the character Tinker Bell, produced by DisneyToon Studios as part of the Disney Fairies franchise after producing a number of direct-to-video Winnie the Pooh films. Voices of Mae Whitman, Lucy Liu, Raven-Symoné, America Ferrera, Kristin Chenoweth and Pamela Adlon are featured in the main films. Six feature films and one TV special were produced: \textit{Tinker Bell}, \textit{Tinker Bell and the Lost Treasure}, \textit{Tinker Bell and the Great Fairy Rescue}, \textit{Secret of the Wings}, \textit{Pixie Hollow Games} (the special), \textit{The Pirate Fairy}, and \textit{Tinker Bell and the Legend of the NeverBeast}. The series is a spin-off of and prequel to \textit{Peter Pan}. Originally developed as a direct-to-video franchise, the series was theatrically released from its third film onwards. \\

\textbf{[5]:} Sir James Matthew Barrie, 1st Baronet, was a novelist and playwright, best remembered as the creator of \textit{Peter Pan}. He was born and educated in Scotland and then moved to London, where he wrote several successful novels and plays. \\
\hline
\end{tabular}
\caption{An example of bad document retrieval using \tRAG (query translated before retrieval). The retrieved documents are relevant to the target query.}
\label{tab:tRAG_wrong}
\end{table*}

\begin{table*}[h!]
\section{Example of \MRAG}
\label{app:MRAG_ok}
\centering
\small
\begin{tabular}{p{16cm}}
\textbf{Question:}\begin{CJK}{UTF8}{gbsn} 英格兰有多少位女王 \end{CJK} \textit{(how many queens have been in England?)} \\
\textbf{Target:} 8 \\ \hline
\textbf{Answer Llama3-8 (no-RAG):} \\ 
\begin{CJK}{UTF8}{gbsn} 实际统治英格兰的女王有四位：玛格丽特一世，玛丽一世，伊丽莎白一世和伊丽莎白二世。女王人数为 4。\end{CJK} \\
\textit{(Four queens actually ruled England: Margaret I, Mary I, Elizabeth I and Elizabeth II. The number of queens is 4.)} \\
\hline
\textbf{Retrieval from from \textbf{W$_{ALL}$}} (Wikipedia in \textbf{ALL} languages):\\

\textbf{[1]:} Since 1 May 1707, when the Kingdom of England and the Kingdom of Scotland were united to form the Kingdom of Great Britain, there have been 13 monarchs of the United Kingdom. The Kingdoms of England and Scotland had been under a commonwealth confederation under the Stuarts since 24 March 1603, and were formally united on 1 May 1707 to form the Kingdom of Great Britain, which was united with the Kingdom of Ireland on 1 January 1801 to form the United Kingdom of Great Britain and Ireland, and the Irish Free State was formed and became independent [...] (\textcolor{blue}{\textit{from \textbf{W$_{Zh}$}})} \\

\textbf{[2]:} This entry lists all female monarchs in the world, including emperors, queens, princes, grand dukes, duchesses, goddesses and other monarchs, and excludes female royals who have ruled on the throne or as regents. In addition, Elizabeth II, who was monarch of the United Kingdom from 1952 to 2022, ruled over 32 Commonwealth realms [...]. (\textcolor{blue}{\textit{from \textbf{W$_{En}$}})} \\

\textbf{[3]:} In less than 200 years, eight queens ruled the Kingdom of England. Among these queens were Matilda, Lady Jane Grey, Mary I, Elizabeth I, Mary II, Anne, Victoria, and Elizabeth II. Despite the tradition that heirs of the throne should be the eldest son of the monarch, the mentioned queens ruled competence and capability [...] (\textcolor{blue}{\textit{from \textbf{W$_{En}$}})} \\

\textbf{[4]:} Empress Matilda, also known as Empress Maud, was one of the claimants to the English throne during the civil war known as the Anarchy. The daughter and heir of Henry I, king of England and ruler of Normandy, she went to Germany as a child when she was married to the future Holy Roman Emperor Henry V. She travelled with the emperor to Italy in 1116, was controversially crowned empress in St Peter’s Basilica [...] (\textcolor{blue}{\textit{from \textbf{W$_{Zh}$}})} \\

\textbf{[5]:} Elizabeth II, known as Elizabeth Alexandra Mary, was Queen of the United Kingdom and other Commonwealth realms from 1952 to 2022 and held the record for the longest reign [...] (\textcolor{blue}{\textit{from \textbf{W$_{Zh}$}})} \\
\hline
\textbf{Answer Llama3-8b (\MRAG):}\\ 
\begin{CJK}{UTF8}{gbsn}
英格兰曾有8位女王作为君主执政....答案是：8 \end{CJK} \\
Thus, England has had 8 queens who ruled as sovereigns.... \textbf{The answer is: 8.} \\
\hline
\end{tabular}
\caption{An example of \MRAG inference for Llama3-8b. To facilitate the reading, we translated retrieved from Chinese Wikipedia (\textcolor{blue}{\textit{ \textbf{W$_{Zh}$}}}) into English (at inference time, we delivered them in Chinese).}
\label{tab:MRAG_ok}
\end{table*}

\begin{table*}[h!]
\section{Example of \MRAG Misleading Answer}
\label{app:MRAG_bad}
\centering
\small
\begin{tabular}{p{16cm}}
\textbf{Question:} \begin{CJK}{UTF8}{mj}
누가 '나는 바비걸' 노래를 만들었나요?
\end{CJK} \textit{(Who made the song i'm a barbie girl?)} \\
\textbf{Target:} \begin{CJK}{UTF8}{mj} ['아쿠아' ,\end{CJK} 'Aqua' ] \\ \hline
\textbf{Answer Llama3-8 (no-RAG):} \\ 
\begin{CJK}{UTF8}{mj} 정답은 Ingemar Örhagen, Matthias Lindblom, 그리고 Candice Posey입니다.\end{CJK} \\
 (The answer is Ingemar Örhagen, Matthias Lindblom and Candice Posey. )  \\
\hline
\textbf{Retrieval from from \textbf{W$_{ALL}$}} (Wikipedia in \textbf{ALL} languages):\\

\textbf{[1]:} The music video for "Barbie Girl" was recorded on August 2005, between 17 and 18. It was directed by Ricardo Vereza, Bidu Madio, Rentz and Mauricio Eça. The video was released on August 30. Kelly plays a determinate and feminist woman, who doesn't need a man [...] (\textcolor{blue}{\textit{from \textbf{W$_{Ko}$}})} \\

\textbf{[2]:} Barbie Girl is a song by the Danish-Norwegian dance-pop group Aqua. It was released in May 1997 as the third single from the group's debut studio album, "Aquarium" (1997). The song was written by Søren Rasted, Claus Norreen, René Dif, and Lene Nystrøm, and was produced by Johnny Jam, Delgado, Rasted, and Norreen. It was written after Rasted [...] (\textcolor{blue}{\textit{from \textbf{W$_{En}$}})} \\

\textbf{[3]:} Barbie and The Three Musketeers is a video premiere animated feature from Universal Pictures, released on DVD on 15 September 2009 in the United States. It is the 16th film in the Barbie animated series and is based on the novel Les Trois Mousquetaires by French writer Alexandre Dumas. The heroine of the film is played by Barbie and her three friends, Teresa, Summer and Nikki, who together realise their dream of becoming  [...] (\textcolor{blue}{\textit{from \textbf{W$_{Ko}$}})} \\

\textbf{[4]:} Barbie as the Princess and the Pauper is a video premiere animated production released by Lionsgate Entertainment on 28 September 2004 on DVD and VHS. It is the fourth film in the Barbie animated series, based on Mark Twain's classic novel The Beggar Prince. The film is also the first animated musical of the Barbie film series, with a total of seven original songs performed by Barbie in two roles a noble princess and a poor village girl.  [...] 
(\textcolor{blue}{\textit{from \textbf{W$_{Ko}$}})}\\

\textbf{[5]:} The song, at first glance, appears to be about the popular children's doll Barbie. However, at second glance, the song contains several sexual overtones, such as "You can brush my hair, undress me everywhere". This caused some controversy upon its release in Denmark, but despite the controversy and perhaps partly due [...] (\textcolor{blue}{\textit{from \textbf{W$_{En}$}})} \\
\hline
\textbf{Answer Llama3-8b (\MRAG):}\\
The seven original songs in Barbie as the Princess and the Pauper are less certain, as they were likely adapted from classical compositions and reworked into the animated film's soundtrack by an unspecified creative team. \textbf{The answer is:} The songs were all composed by Wolfgang Amadeus Mozart in collaboration with contemporary pop producers. (\textcolor{blue}{\textit{translated from Korean}})\\
\textbf{Answer Llama3-8b (\cRAG):}\\
The song "Barbie Girl" was performed by the Danish-Norwegian [...] \textbf{The answer is: Aqua.} (\textcolor{blue}{\textit{translated from Korean}}) \\
\textbf{Answer GPT-4o (\MRAG):}\\
The song "Barbie Girl" was written by Søren Rasted, Claus Norreen, René Dif, and Lene Nystrøm, members of the Danish-Norwegian dance-pop group Aqua. \textbf{The answer is: Aqua.} (\textcolor{blue}{\textit{translated from Korean}})
 \\
\hline
\end{tabular}
\caption{An example of \MRAG inference. We have translated documents and answers into English to facilitate the understanding as in Table \ref{tab:MRAG_ok}.}
\label{tab:MRAG_bad}
\end{table*}

\end{document}